\documentclass[10pt]{scrartcl}

\usepackage{makeidx}
\usepackage{graphicx}
\usepackage{caption}
\usepackage{subcaption}
\usepackage{amssymb}
\usepackage{amsmath}
\usepackage{amscd}
\usepackage{tikz}
\usetikzlibrary{matrix,arrows}
\usepackage{multicol}
\usepackage{setspace}
\usepackage{algorithm}
\usepackage[noend]{algpseudocode}
\usepackage{url}
\usepackage{color}

\newtheorem{definition}{Definition}[section]
\newtheorem{theorem}[definition]{Theorem}
\newtheorem{corollary}[definition]{Corollary}
\newtheorem{proposition}[definition]{Proposition}

\newcommand{\qed}{\hspace*{\fill} $\blacksquare$}
\newcommand{\proof}{Proof:\ }

\newcommand{\commentout}[1]{}

\newcommand{\R}{\mathbb{R}}                    

\newcommand{\abs}[1]{\mathop{\left\lvert #1 \right\rvert}} 
\newcommand{\args}[1]{\mathop{\left( #1 \right)}} 
\newcommand{\inner}[1]{\mathop{\left\langle #1 \right\rangle}}
\newcommand{\norm}[1]{\mathop{\left\lVert #1 \right\rVert}}
\newcommand{\cbrace}[1]{\mathop{\left\{ #1 \right\}}}
\newcommand{\bracket}[1]{\mathop{\left[ #1 \right]}}

\newcommand{\normS}[2]{\mathop{\left\lVert #1 \right\rVert#2}}

\renewcommand{\S}[1]{{\mathcal{#1}}}           	
\def\vec#1{\mathchoice{\mbox{\boldmath$\displaystyle#1$}}
{\mbox{\boldmath$\textstyle#1$}}
{\mbox{\boldmath$\scriptstyle#1$}}
{\mbox{\boldmath$\scriptscriptstyle#1$}}}

\renewenvironment{cases}{%
\left\{\begin{array}{c@{\quad : \quad}l}}%
{%
\end{array}\right.}

\newcounter{part_counter}
\begin{document}
\nocite{*}
\title{Properties of the Sample Mean in Graph Spaces and the Majorize-Minimize-Mean Algorithm}
\author{Brijnesh J.~Jain \\
 Technische Universit\"at Berlin, Germany\\
 e-mail: brijnesh.jain@gmail.com}
 
\date{}
\maketitle

\begin{abstract} 
%
%
One of the most fundamental concepts in statistics  is the concept of sample mean. Properties of the sample mean that are well-defined in Euclidean spaces become unwieldy or even unclear in graph spaces. Open problems related to the sample mean of graphs include: non-existence, non-uniqueness, statistical inconsistency, lack of convergence results of mean algorithms, non-existence of midpoints, and disparity to midpoints. We present conditions to resolve all six problems and propose a Majorize-Minimize-Mean (MMM) Algorithm. Experiments on graph datasets representing images and molecules show that the MMM-Algorithm best approximates a sample mean of graphs compared to six other mean algorithms.
\end{abstract} 

\newpage

\tableofcontents

\newpage

\section{Introduction}

Statistical inference deduces properties about a population by analyzing a subset of sampled data. One central path departs from the fundamental concept of mean, leads via the normal distribution and the Central Limit Theorem to statistical estimation using the maximum likelihood method. The maximum likelihood method in turn is a fundamental approach that provides probabilistic interpretations to learning methods from pattern recognition.

This central path is well-defined in Euclidean spaces, but becomes unclear in mathematically less structured spaces. Since an increasing amount of non-Euclidean data is being collected and analyzed in ways that have not been realized before, statistics is undergoing an evolution \cite{Kim2010}. Examples of this evolution are contributions to statistical analysis of shapes \cite{Bhattacharya2012,Dryden1998,Huckemann2010,Kendall1984}, complex objects \cite{Marron2014,Wang2007}, and tree-structured data \cite{Feragen2011a,Feragen2011b,Feragen2013,Wang2007}. 

The focus of this contribution is on statistical analysis of graphs, which is less explored than, for example, the space of shapes. Graphs occur in different areas such as computer vision, network analysis, chemo- and bioinformatics \cite{Conte2004,Foggia2014,Gao2010,Livi2013}. The problem of adopting statistical techniques for analyzing graphs is referred to as the gap between structural and statistical pattern recognition \cite{Bunke2001,Bunke2012,Duin2012}. 

The first step towards a statistical analysis of graphs was the pioneering work on the sample mean by Jiang et al.~\cite{Jiang2001}. A sample mean of $n$ graphs $X_1, \ldots. X_n$ is any graph that minimizes the sample Fr\'echet function \cite{Frechet1948}
\begin{align*}
F_n(X) = \sum_{i=1}^n d\!\args{X_i, X},
\end{align*}
where $d(X, Y)$ is a graph edit distance. Attention on the sample mean of graphs predominantly focused on devising efficient algorithms for minimizing $F_n(X)$ as in \cite{Ferrer2010a, Hlaoui2006, Mukherjee2009}. But there are hardly any studies that aim at understanding the statistical and geometrical properties of the sample mean as in \cite{Ginestet2012,Jain2008}. Due to the lack of comprehensive theoretical studies, we face fundamental problems related to the sample mean of graphs:
\begin{enumerate}
\itemsep0em
\item[(P1)] A mean may not exist.
\item[(P2)] A mean may not be unique. 
\item[(P3)] A sample mean may not be a consistent estimator of a population mean.
\item[(P4)] Solutions of mean algorithms may not satisfy necessary conditions of optimality. 
\item[(P5)] Two graphs may not have a midpoint. 
\item[(P6)] A mean of two graphs may not be a midpoint.
\end{enumerate}
In addition to the theoretical problems, the sample mean of graphs suffers from computational issues: (C1) the function $F_n(X)$ is non-convex and may have many local minima, and (C2) computation of a graph edit distance is NP-hard. 

The consequences of problems (P1)--(P6) and (C1)--(C2) are manifold: First, it is unclear whether the concept of sample mean of graphs is useful to describe a population of graphs. Second, without understanding the concept of mean, we will not understand adaptions of other concepts and results from traditional statistics to the domain of graphs. Examples include the normal distribution and the Central Limit Theorem. This in turn means that the "bread-and-butter" tools of statistics remain inaccessible for narrowing the gap between structural and statistical pattern recognition. Third, the above stated problems propagate to further problems of learning methods such as prototype-based clustering \cite{Gold1996b,Gunter2002} and principal component analysis. 

We do not understand graph spaces well, because the graph edit distance is not a metric. Results from geometry show that being a metric space is a necessary (but not sufficient) condition to eliminate all problems (P1)--(P6). To establish a theory of statistical analysis on graphs, we need to restrict the representative flexibility of the graph edit distance in favor of analytical flexibility. An example of an analytically more well-behaved graph edit distance is the graph edit kernel metric \cite{Jain2015}. The graph edit kernel metric is not an artificial construction for the sole purpose of analyzing graphs, but rather is a common and widely applied (dis)similarity function on graphs 
\cite{Almohamad1993,Caetano2007,Cour2006,Gold1996,Leordeanu2005,Leordeanu2009,Umeyama1988,Wyk2002,Zaslavskiy2009,Zhou2012}.

This paper aims at providing a first step to establish a theory of statistical graph space analysis. The contributions fall into two areas: 
\begin{enumerate}
\item \emph{Theory}: We propose conditions that resolve problems (P1)--(P6) in graph spaces endowed with the graph edit kernel metric. The proposed approach combines results on sample Fr\'echet means in metric spaces \cite{Bhattacharya2012} with results from the geometry of graph edit kernel spaces \cite{Jain2015}. 
\item \emph{Algorithms}: We propose a mean algorithm that belongs to the class of Majorize-Minimize Algorithms \cite{Hunter2004}. This class of algorithms includes the EM algorithm as a special case and provides access to general convergence results. In experiments, we compared the Majorize-Minimize-Mean (MMM) Algorithm against six common related variants reported in the literature.
\end{enumerate}

The rest of this paper is structured as follows: After discussing related work, Section \ref{sec:background} introduces attributed graphs and the graph edit kernel metric. In Section \ref{sec:Theory}, we develop the theory of sample mean in graph edit kernel spaces on the basis of Fr\'echet functions and presents the MMM-Algorithm. In Section \ref{sec:Experiments}, we assess the performance of the MMM-Algorithm and discuss the results. Finally, Section \ref{sec:Conclusion} concludes with a summary of the main result and an outlook for further research. Proofs are delegated to the Appendix.

\subsection{Related Work}

\paragraph*{Theory.} Research on the sample mean of graphs based on Fr\'echet's formulation \cite{Frechet1948} started with the pioneering work of Jiang et al.~\cite{Jiang2001}. They used the general graph edit distance and studied two types of minimizers of the sample Fr\'echet function $F_n(X)$. The first type of minimizers is the medoid (set median) that picks the minimum of $F_n(X)$ from the $n$ sample graphs and the second type is the sample median (generalized median) that minimizes $F_n(X)$ over the entire graph space. In the following we refer to all measures of central tendency as sample mean for the sake of convenience.

The work by Jiang et al.~simultaneously triggered two directions of research. The first direction focused on devising algorithms for minimizing different formulations of the sample Fr\'echet function \cite{Bardaji2010, Ferrer2008, Ferrer2009a, Ferrer2009b, Ferrer2010a, Ferrer2010b, Hlaoui2006, Mukherjee2009, Raveaux2011}. The second direction developed prototype-based clustering algorithms partly resulting in novel ways to compute a sample mean \cite{Bonev2007, Bunke2003, Chen2012, Ferrer2009c,Gunter2002,Seeland2010}. 

All these studies did neither consider fundamental statistical nor geometrical properties of the sample mean. From the beginning, it was well-known hat a sample mean is not unique (P2). But it was unclear, whether a sample mean exists (P1) and whether a mean algorithm converges to a solution (P4). Furthermore, it was unclear wether a sample mean converges to a population mean (if it exists) when the number of sample graphs tends to infinity. Sample mean approaches \cite{Ferrer2008,Ferrer2010a,Ferrer2011} via dissimilarity representations \cite{Pekalska2005} compounded the consistency problem. 

In \cite{Jain2010}, we connected the sample mean of graphs to the theory of Fr\'echet functions for which mathematical statistics provides deep results \cite{Bhattacharya2012}. We proposed statistical consistency without presenting a proof. Since then little progress had been made in understanding the sample mean of graphs. As a consequence, statistical analysis of graphs is less developed than statistical analysis of other data structures such as shapes \cite{Dryden1998,Huckemann2010,Kendall1984}, complex objects \cite{Marron2014,Wang2007}, and tree-structured data \cite{Feragen2011a,Feragen2011b,Feragen2013,Wang2007}. The geometric part of this contribution  is inspired by Feragen et al.~\cite{Feragen2013}.

\paragraph*{Algorithms.} Different types of algorithms were proposed to approximate a sample mean. Examples include genetic algorithms \cite{Jiang2001,Ferrer2009a,Raveaux2011}, greedy algorithms \cite{Hlaoui2006}, graph spectral methods \cite{Ferrer2006, White2006},  and fusion methods based on optimal alignments \cite{Bonev2007,Chen2012,Gold1996b,Jain2004,Jain2008,Jain2009,Jain2011b,Jain2012,Jain2013,Lozano2003,Lozano2006,Serratosa2002,Serratosa2003}. The majority of fusion methods process the sample graphs incrementally and loop only once through the sample. The performance of incremental fusion methods depend on the initialization of the reference graph and on the order of how the graph are presented. 

Most fusion approaches lack convergence and consistency statements. An exception is an incremental fusion method proposed in prior work \cite{Jain2009} that belongs to the class of stochastic generalized gradient (SGG) algorithms \cite{Ermoliev1998,Norkin1986}. Theoretically, the field of stochastic optimization provides a different way to resolve consistency and convergence issues. Statistical consistency established by stochastic optimization is only valid for the class of stochastic generalized gradient methods, whereas statistical consistency proposed in this work is independent of the algorithm and only depends on the global minima of the sample Fr\'echet function.

In prior work \cite{Jain2009}, we empirically studied an algorithm closely related to the proposed MMM-Algorithm. We studied a batch version of the SGG-Algorithm without presenting a convergence result. In contrast to MMM, batch SGG uses a step size parameter.  Results of batch SGG were not convincing, because we selected sub-optimal values for the step-size parameter. However, the key difference between batch SGG and the MMM-Algorithm is not the choice of step size parameter but the underlying concepts from which both algorithms are derived. The underlying concept of batch SGG is the concept of generalized gradient and the underlying concept of MMM is the necessary condition of optimality proposed in Theorem \ref{theorem:nesuco}. Though both concepts result into almost the same algorithms, they have a significant impact in theory and practice. Theoretically, we can prove convergence by invoking Zangwill's Theorem \cite{Zangwill1969}. Practically, we get rid of the step size parameter and thereby gain increase in solution quality compared to the results in \cite{Jain2009}. 

Fusion methods for mean computation have been piled up without a unifying theory that provides a mathematical justification and place them in proper context. As unifying scheme, we suggest Theorem \ref{theorem:nesuco} on necessary conditions of optimality, which we proved in prior work \cite{Jain2008}, Theorem 3. We include this result due to its pivotal importance and for the sake of coherence and completeness. The importance of Theorem \ref{theorem:nesuco} result from its multiple applications. For example, from Theorem \ref{theorem:nesuco} follows that the problem of minimizing a Fr\'echet sample function is equivalent to the multiple alignment (labelling) problem \cite{Jain2008,Rebagliati2012}. This result brings together algorithms from two different domains, ordinary sample mean computation from graph-based representation and fusion schemes from progressive alignment for multiple sequence alignment \cite{Saitou1987}, which can all be placed under the common umbrella of Theorem \ref{theorem:nesuco}.

\section{Graph Edit Kernel Spaces}\label{sec:background}

This section describes the space of graphs endowed with a geometric version of the graph edit distance.

\paragraph{Attributed Graphs.} Let $\S{A} = \R^d$ be the set of node and edge attributes. An attributed graph is a triple $X = (\S{V}, \S{E}, \alpha)$, where $\S{V}$ represents a finite set of nodes, $\S{E} \subseteq \S{V} \times \S{V}$ a set of edges, and $\alpha: \S{V} \times \S{V} \rightarrow \S{A}$ is an attribute function satisfying
\begin{enumerate}
\item $\alpha(i, j) \neq \vec{0}$ for all edges $(i, j) \in \S{E}$ 
\item $\alpha(i, j) = \vec{0}$ for all non-edges $(i, j) \notin \S{E}$ 
\end{enumerate}
where $i, j \in \S{V}$ are distinct nodes. According to the above definition, graphs have the following properties:
\begin{enumerate}
\item Attributes $\alpha(i,i)$ of nodes $i \in \S{V}$ may take any value from $\S{A}$.
\item Graphs are complete by regarding non-edges as edges with zero attribute $\vec{0}$. 
\item The definition comprises directed as well as undirected graphs.
\end{enumerate}

\paragraph*{Matrix Representations.}
It is convenient to identify graphs with sets of matrices. A graph $X$ is completely specified by a matrix representation $\vec{X} = (\vec{x}_{ij})$, where the elements $\vec{x}_{ij} \in \S{A}$ represent the node and edges attributes for all $i,j \in \S{V}_X$.

The particular form of a matrix representation of a graph depends on how we order its nodes. Permuting the order of the nodes of a graph may result in a different matrix representation. By permuting the nodes in all possible ways, we obtain the equivalence class $[\vec{X}]$ of all matrix representations of $X$. Thus, we can identify graphs $X$ with sets of matrices that represent $X$. We therefore write $\vec{X} \in X$ if $\vec{X}$ represents $X$.

\paragraph*{Graph Edit Kernels.}

Next, we endow the set $\S{G_A}$ with a similarity function, called graph edit kernel. We say two graphs $X$ and $Y$ are size-aligned if the smaller of both graphs is expanded to size $n = \max\args{\abs{\S{V}_X}, \abs{\S{V}_Y}}$ by adding isolated nodes with zero attribute $\vec{0}$. By $\widetilde{X}$ and $\widetilde{Y}$ we denote the size-aligned graphs of $X$ and $Y$. Then the graph edit kernel is defined by
\[
\kappa: \S{G_A} \times \S{G_A} \rightarrow \R, \quad (X, Y) \mapsto \max \cbrace{\inner{\vec{X},\vec{Y}} \,:\, \vec{X} \in \widetilde{X}, \vec{Y} \in \widetilde{Y}},
\]
where 
\[
\inner{\vec{X},\vec{Y}} = \sum_{i,j} \vec{x}_{ij}^T \vec{y}_{ij}
\]
is the inner product between two matrices with elements from $\S{A}$.\footnote{More generally, we can replace the inner product $ \vec{x}_{ij}^T \vec{y}_{ij}$ by a positive definite kernel  $k( \vec{x}_{ij}, \vec{y}_{ij})$.}  A graph edit kernel induces a metric $\delta: \S{G_A} \times \S{G_A} \rightarrow \R$ defined by
\begin{align*}
\delta(X, Y) = \sqrt{\kappa(X, X) + \kappa(Y, Y) - 2\kappa(X, Y)}
\end{align*}
for all $X, Y \in \S{G_A}$.  A graph edit kernel space is a graph edit distance space $(\S{G_A}, \delta)$, where $\delta$ is a metric induced by a graph edit kernel. 

\begin{figure}[t]
\centering
\includegraphics[width=0.7\textwidth]{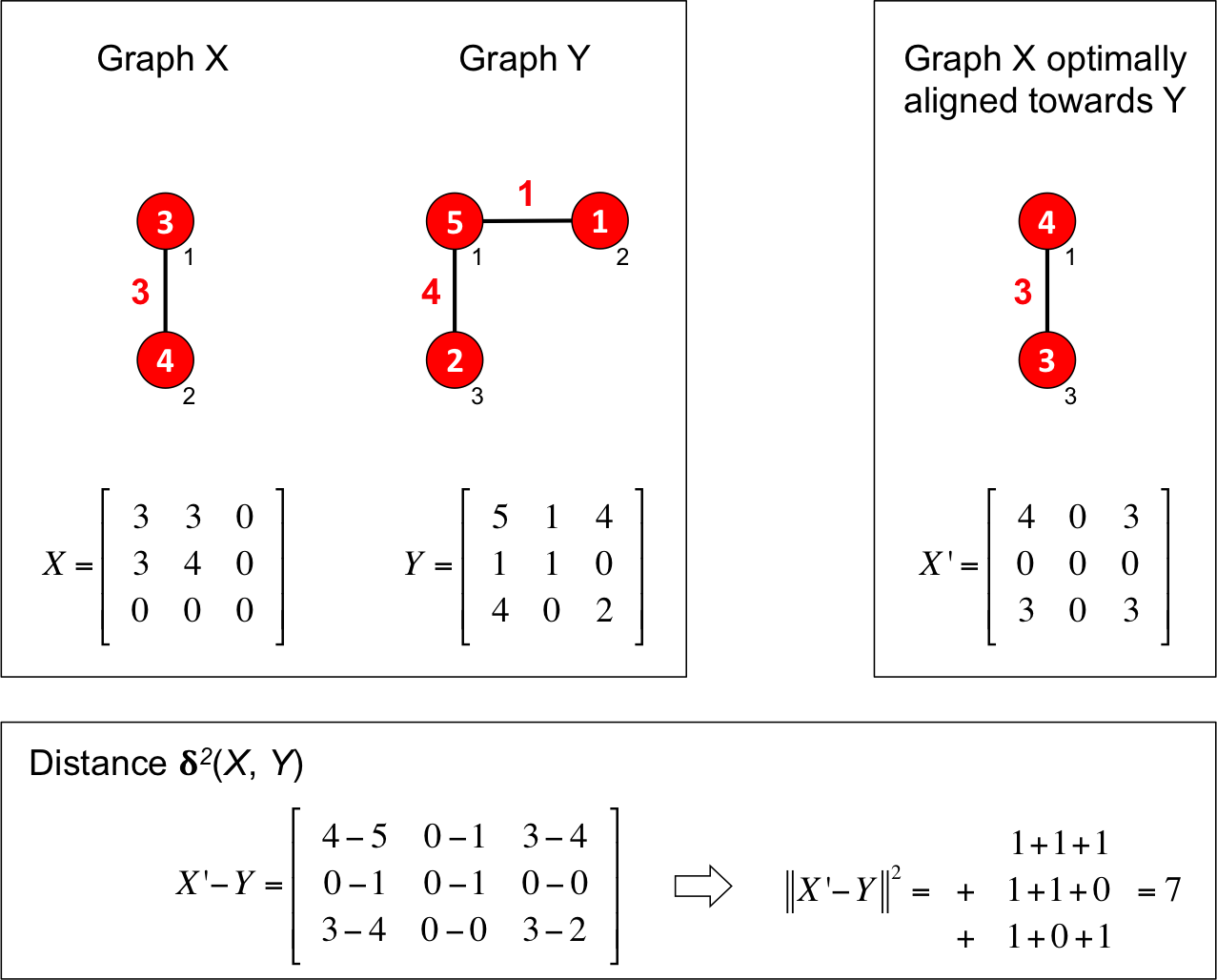}
\caption{\small The upper left box shows two attributed graphs $X$ and $Y$, where attributes are weights. White numbers inside the nodes are the node attributes and red numbers attached to the edges are the edge attributes. Small black numbers next to the nodes are their unique identifiers. These identifiers correspond to the order of nodes according to the matrix representation given below of each graph. In this example, we assume that all graphs are of bounded order $3$. Then all matrix representations have dimension $3 \times 3$. The matrix representation of the two-node graph $X$ has a padding zero column and row. The box at the upper right shows the identifiers of the nodes of graph $X$ when optimally aligned towards graph $Y$ and its matrix representation $\vec{X}'$. The graph edit kernel metric is defined by $\delta(X, Y) = \norm{\vec{X}' - \vec{Y}} = \sqrt{7}$.} 
\label{fig:optimal_alignment}
\end{figure}

We can equivalently express the metric $\delta$ by
\begin{align*}
\delta(X, Y) &= \min \cbrace{\norm{\vec{X}-\vec{Y}} \,:\, \vec{X} \in \widetilde{X}, \vec{Y} \in \widetilde{Y}}\\
&= \min \cbrace{\norm{\vec{X}-\vec{Y}} \,:\, \vec{Y} \in \widetilde{Y}}\\
&= \min \cbrace{\norm{\vec{X}-\vec{Y}} \,:\, \vec{X} \in \widetilde{X}},
\end{align*}
where $\vec{X} \in \widetilde{X}$ in the second equation and $\vec{Y} \in \widetilde{Y}$ in the third equation are arbitrarily chosen representations.

We say, a matrix representation $\vec{X}$ is optimally aligned towards matrix representation $\vec{Y}$ if $\delta(X, Y) = \norm{\vec{X}-\vec{Y}}$. Occasionally, we say a graph $X$ is optimally aligned to graph $Y$ for the sake of convenience, but always mean their respective representations under consideration.

\paragraph*{Assumption.} For the sake of mathematical convenience, we assume that graphs have bounded number $n$ of nodes. Graphs of order $m < n$ can be extended to graphs of order $n$ by adding $n - m$ isolated vertices with zero attribute. Note that this assumption is purely technical without computational and limiting impact in practice. In a practical setting, we neither extend graphs to order $n$ nor even know the exact value of $n$.
\qed

\bigskip

Figure \ref{fig:optimal_alignment} illustrates the concepts of graph edit kernel metric, optimal alignment, and graph extension to a bounded order.

\section{Fr\'echet Functions on Graphs}\label{sec:Theory}

This section studies properties of the concept of sample mean in graph edit kernel spaces. We provide conditions to resolve problems (P1)--(P6) and propose a Majorize-Minimize-Mean (MMM) Algorithm.

\subsection{Fr\'{e}chet Functions in Euclidean Spaces}
The goal of this section is to describe the basic idea of Fr\'{e}chet functions for generalizing statistical measures of central tendency to arbitrary distance spaces.

\medskip

The standard definition of a sample mean $\vec{\mu}_n$ of $n$ data points $\vec{x}_1, \vec{x}_2, \ldots, \vec{x}_n \in \R^d$ is of the form
\begin{align*}
\vec{\mu}_n = \frac{1}{n} \sum_{i=1}^n \vec{x}_i.
\end{align*}
This definition requires a well-defined vector addition, which is not available in arbitrary distance spaces. To generalize the concept of sample mean to distance spaces, we resort to an equivalent formulation of the sample mean suggested by Fr\'echet \cite{Frechet1948}. The main idea is based on the fact that the sample mean $\vec{\mu}_n$ is the unique minimum of the \emph{sample Fr\'echet function} 
\begin{align*}
F_n(\vec{z}) = \sum_{i=1}^n \normS{\vec{x}_i-\vec{z}}{^2}.
\end{align*}

In statistics the sample mean is used to estimate the population mean. As a result of the Law of Large Numbers, the sample mean is a consistent estimator of the population mean
\begin{align*}
\vec{\mu} = \int \vec{x} dp(\vec{x}),
\end{align*}
 where $p(\vec{x})$ is a probability distribution on $\R^d$. The population mean $\vec{\mu}$ minimizes the Fr\'echet function 
\begin{align*}
F(\vec{z}) = \int_{\R^d} \normS{\vec{x} - \vec{z}}{^2} dp(\vec{x}).
\end{align*}
The Fr\'echet function $F(\vec{z})$ is more useful for generalizing the concept of population mean to distance spaces than the standard definition of the population mean.

Fr\'echet's idea has been generalized to other measures of central tendency. For example, a sample median of $n$ data points $\vec{X}_{\!i} \in \R^d$ minimizes the sample Fr\'echet function 
\begin{align*}
F_n(\vec{z}) = \sum_{i=1}^n \norm{\vec{x}_i-\vec{z}}.
\end{align*}
Similarly, the population median minimizes the Fr\'echet function 
\begin{align*}
F(\vec{z}) = \int \norm{\vec{x}-\vec{z}} dp(\vec{x}).
\end{align*}
In general, we can study measures of central tendencies as minimizers of a sample Fr\'echet function of the general form
\[
F_n(\vec{z}) = \sum_{i=1}^n L\args{\norm{\vec{x}_i-\vec{z}}},
\]
where $L:\R \rightarrow [0, \infty)$ is a continuous loss function. The identity loss $L(a) = a$ refers to the median formulation and the squared loss $L(a)=a^2$ to the mean formulation.

\subsection{Fr\'{e}chet Functions in Graph Edit Kernel Spaces}
Using Fr\'echet's optimization-based formulations, we generalize the concept of sample mean to measures of central tendencies in graph edit kernel spaces. For the sake of convenience, we subsume measures of central tendency under the common term \emph{sample mean} and (\emph{population}) \emph{mean}.

\bigskip

Let $Q$ be a probability distribution on the graph edit kernel space $\args{\S{G_A}, \delta}$.\footnote{More precisely: Let $\args{\Omega, \S{A}, P}$ be an abstract probability space. A random element $X$ on $\S{G_A}$ is a mapping $X: \Omega \rightarrow \S{G_A}$ measurable with respect to the Borel $\sigma$-algebra $\S{B}$ on $\S{G_A}$ induced by the metric $\delta$. Then $Q = P \circ X^{-1}$ is a probability measure on the measurable space $\args{\S{G_A}, \S{B}}$.} Suppose that $\S{S}_n = \args{X_1, X_2, \ldots, X_n}$ is a sample of $n$ graphs from $\S{G_A}$ drawn i.i.d.~according to the probability distribution $Q$. Then the \emph{sample Fr\'{e}chet function} of $\S{S}_n$ is of the form
\begin{align*}
F_n: \S{G_A} \rightarrow \R, \quad Z \mapsto \sum_{i=1}^n L\args{\delta(X_i, Z)}
\end{align*}
where $L:\R \rightarrow [0, \infty)$ is a continuous loss function. 

In general, a minimum of the sample Fr\'echet function in arbitrary distance spaces may not exist  (P1). If a minimum exists, it is not necessarily unique  (P2). The sample Fr\'echet function is not differentiable unless the underlying space is a Euclidean or Banach space. Thus, it is unclear how to minimize a sample Fr\'echet function efficiently and how to state necessary conditions of optimality  (P4). To cope with these issues we define the \emph{sample mean set} of $\S{S}_n$ as 
\[
\S{F}_n = \cbrace{M \in \S{G_A} \,:\, M = \arg\min_Z F_n(Z)}.
\]
The elements of $\S{F}_n$ are the \emph{sample means} of $\S{S}_n$. The minimum value of $F_n$ on $\S{G_A}$ is called the \emph{sample variation} of $\S{S}_n$ and is denoted by $V_n$.

To generalize the statistical concept of population mean, we introduce the \emph{Fr\'{e}chet function} of probability distribution $Q$ defined as
\[
F: \S{G_A} \rightarrow \R, \quad Z \mapsto \int_{\S{G_A}} L\args{\delta(X, Z)}\, dQ(X).
\]
The set 
\[
\S{F}_Q = \cbrace{M \in \S{G_A} \,:\, M = \arg\min_Z F(Z)}
\]
is the (\emph{population}) \emph{mean set} of $Q$ and its elements are the (\emph{population}) \emph{means} of $Q$. The minimum value of $F$ on $\S{G_A}$ is called the (\emph{population}) \emph{variation} of $Q$ and is denoted by $V_Q$. 

In arbitrary distance spaces, the Fr\'echet function may suffer from similar anomalies as the sample Fr\'echet function. In addition, a statistical justification of a sample mean is unclear for distance spaces (P3). It can happen that a sample mean systematically misestimates a population mean or even tries to estimate something that does not exist.

\subsection{Necessary Conditions of Optimality}

The first result presents necessary conditions of optimality for the sample Fr\'{e}chet function with squared loss. The result is not new and has been proved in prior work \cite{Jain2008}, Theorem 3.  We include this result and a revised proof due to its importance and for the sake of coherence and completeness.

\begin{theorem}\label{theorem:nesuco}
Consider the sample Fr\'{e}chet function 
\[
F_n(Z) = \sum_{i=1}^n \delta^2\args{X_i, Z},
\]
of $n$ graphs $X_1, \ldots, X_n \in \S{G_A}$. Every representation $\vec{M}$ of a local minimum $M \in \S{G_A}$ of $F_n(Z)$ is of the form
\begin{align}\label{eq:nesuco}
\vec{M} = \frac{1}{n} \sum_{i=1}^n \vec{X}_i,
\end{align}
where $\vec{X}_{\!i} \in X_i$ are representations optimally aligned with $\vec{M}$.
\end{theorem}

\noindent
\proof Section \ref{app:proof-theorem:nesuco}. \qed

\begin{figure}[t]
\centering
\includegraphics[width=0.75\textwidth]{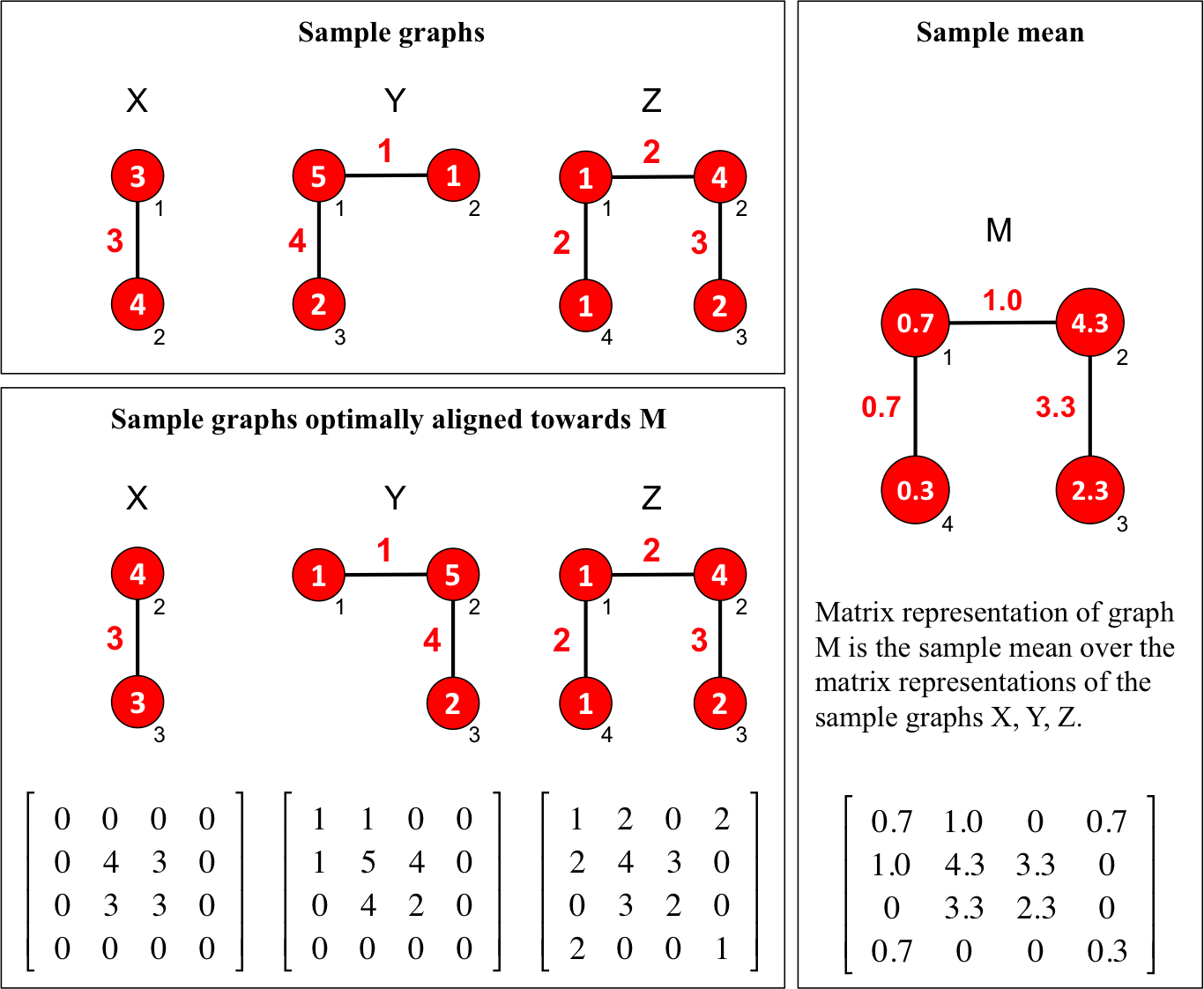}
\caption{\small The upper left box shows three sample graphs $X$, $Y$, and $Z$. The right box show the sample mean $M$ of $X$, $Y$, and $Z$ with rounded attribute values. The lower right box shows the sample graphs optimally aligned to the sample mean $M$ and their matrix representations. The matrix representation of the mean $M$ is the mean of the matrix representations of the aligned sample graphs.} 
\label{fig:example_mean_graph}
\end{figure}

\bigskip

Figure \ref{fig:example_mean_graph} shows an example of a sample mean $M$ of three graphs and shows the form of $M$ according to Equation \eqref{eq:nesuco}. In addition to describing the form of a sample mean, Theorem \ref{theorem:nesuco} has the following implications: 
\begin{enumerate}
\item 
It gives rise to the MMM-Algorithm proposed later in this section. 
\item 
It places a number of fusion-based mean algorithms (see Section \ref{app:sec:algorithms}) in proper context and provides a unifying scheme. 
\item It shows that minimizing the sample Fr\'echet function over the uncountable graph space $\S{G_A}$ is equivalent to minimizing the sample Fr\'echet function over the finite set of all matrix representations that can be obtained by Equation \eqref{eq:nesuco}. For this, we consider all possible combinations of selecting a matrix representation of each sample graphs and take the average as in Equation \eqref{eq:nesuco}. The average with minimum sample variation represents a sample mean. Unfortunately, the number of different representations of a graph increases exponentially with the number of nodes. However, this insight gives rise to an interesting result shown next. 
\item It gives rise to the equivalence between the problem of minimizing a sample Fr\'echet function and the problem of multiple alignment
\[
\args{\vec{X}_1^*, \ldots, \vec{X}_n^*} = \arg\min \cbrace{ \sum_{i = 1}^n \sum_{j = i+1}^n \delta\args{X_i, X_j} \,:\, \vec{X}_1 \in X_1, \ldots, \vec{X}_n \in X_n}.
\]
This equivalence has been shown in \cite{Jain2008,Rebagliati2012}.
\end{enumerate}

\subsection{Existence and Consistency}

We show that a mean of graphs exists and that a sample mean is a consistent estimator of a mean in the sense of Bhattacharya-Patrangenaru \cite{Bhattacharya2003}. In addition, we show that the sample variation is a strongly consistent estimator of the variation. As a consequence, it is unlikely that we systematically misestimate a population parameter or even try to estimate something that does not exist.

\bigskip

The sample mean set $\S{F}_n$ is a Bhattacharya-Patrangenaru (BP) strongly consistent estimator of the mean set $\S{F}_Q$, if $\S{F}_Q \neq \emptyset$ and if for every $\varepsilon > 0$ and for almost every $\omega \in \Omega$ there is an integer $N = N(\varepsilon, \omega)$ such that 
\[
\S{F}_n \subseteq \S{F}_Q^\varepsilon = \cbrace{Z \in \S{G_A} \,:\, \delta\args{Z, \S{F}_Q} \leq \varepsilon}
\]
for all $n \geq N$.

\begin{theorem}
Suppose that the Fr\'{e}chet function $F(Z)$ of $Q$ corresponding to a loss function $L(a)= a^p$ with $p \geq 1$ is finite on $\S{G_A}$. Then the following holds:
\begin{enumerate}
\item 
$\S{F}_Q$ is non-empty and compact.
\item 
$\S{F}_n$ is a BP strongly consistent estimator of  $\S{F}_Q$.
\item 
$V_n$ is a strongly consistent estimator of $V_Q$.
\end{enumerate}
\end{theorem}

\noindent
\proof
The graph edit kernel space is a metric space \cite{Jain2015}, Prop.~3.8. Every closed bounded subset of $\S{G_A}$ is compact \cite{Jain2015}, Theorem 3.3. Then the assertions follow from \cite{Bhattacharya2012}, Theorem 2.3 and Prop.~2.8.
\qed

\subsection{Sufficient Conditions for Uniqueness}

We present sufficient conditions under which a population mean and a sample mean are uniquely determined. For this, we adopt some geometric figures from the Euclidean space and combine them with graph-theoretic concepts.

\paragraph*{Balls, Rays, and Cones.} 

\begin{figure}[t]
\centering
\includegraphics[width=0.5\textwidth]{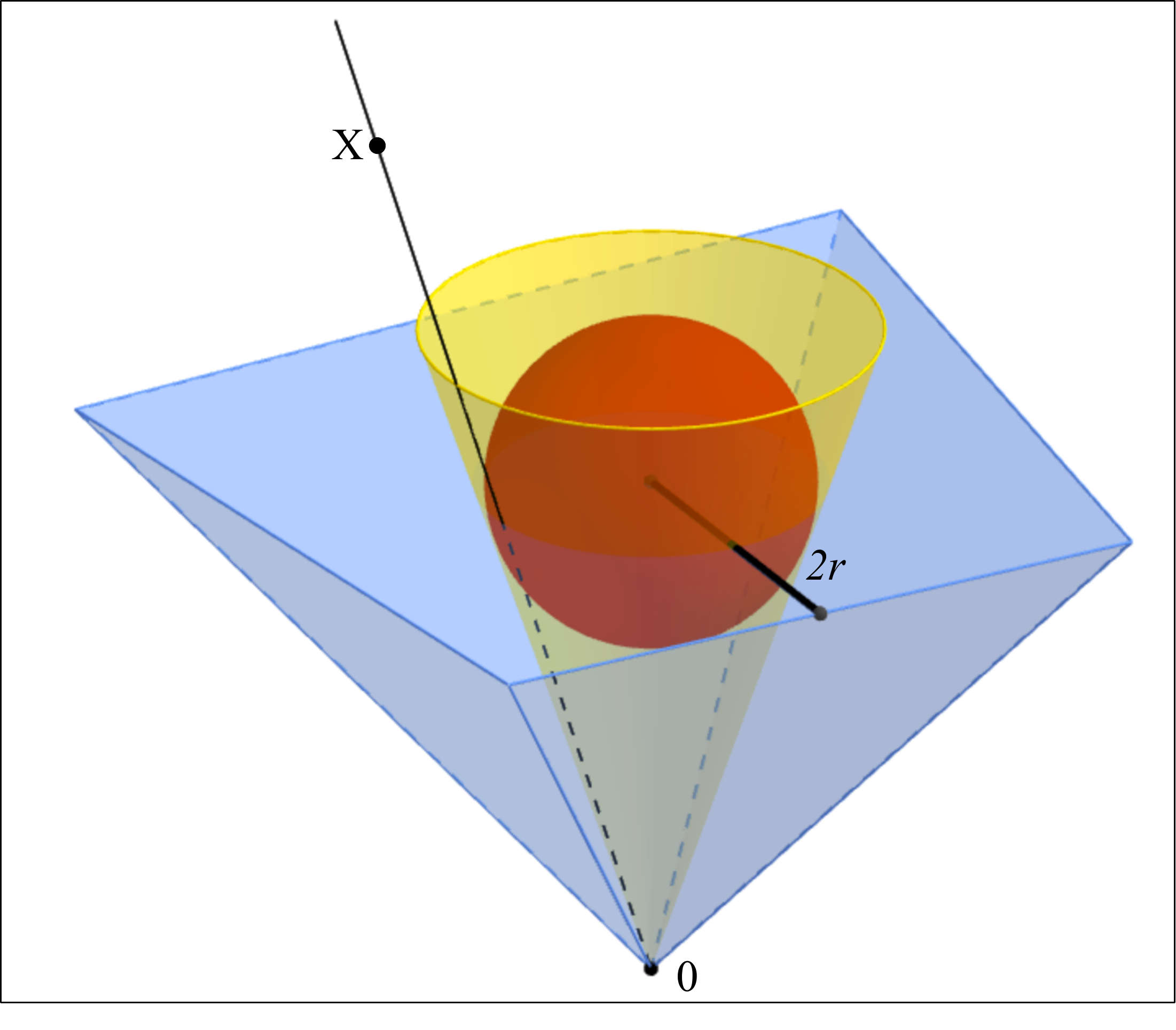}
\caption{\small Geometric visualization of a cone $\S{C}(Z,r)$ circumscribing a ball $\S{B}(Z,r)$ with center $Z$ and radius $r$. The cone $\S{C}(Z,r)$ is shown in yellow and the ball is shown in red. A graph $X$ is an element of $\S{C}(Z,r)$ if there is a non-negative scalar $\lambda \geq 0$ such that $\lambda X$ lies in the ball $\S{B}(Z,r)$. The blue polyhedral cone $\S{D}_Z$ depicts the graph space from the point of view of $Z$. The cone $\S{D}_Z$ with apex $0$ is obtained by the intersection of finitely many half-spaces. The shortest distance of $Z$ from the boundary of $\S{D}_Z$ is $2r$. } 
\label{fig:cone}
\end{figure}

The ball $\S{B}(Z, r)$ with center graph $Z$ and radius $r$ is defined by the set 
\[
\S{B}(Z, r) = \cbrace{X \in \S{G_A} \,:\, \delta(X, Z) \leq r}. 
\]
The scalar multiplication of a scalar $\lambda \in \R$ with a graph $X \in \S{G_A}$ is a graph $Y = \lambda X$ obtained by multiplying all node and edge attributes of $X$ with $\lambda$. As shown in \cite{Jain2015}, we have 
\begin{align*}
\delta(\lambda X, Y) &= \lambda \delta(X, Y)
\end{align*}
for all non-negative scalars $\lambda \geq 0$ and for all $X, Y \in \S{G_A}$. The ray $\S{R}_X$ emanating from the zero graph $0$ and passing through graph $X$ is a set of the form
\[
\S{R}_X = \cbrace{\lambda X \,:\, \lambda \geq 0}.
\]
A pointed cone $\S{C}$ with apex $0$ is a union of rays emanating from the zero graph such that $\S{C}$ s a convex set. A cone $\S{C}(Z, r)$ circumscribing a ball $\S{B}(Z, r)$ is the union of all rays $\S{R}_X$ that have non-empty intersection with $\S{B}(Z, r)$, that is 
\[
\S{C}(Z,r) = \cbrace{X \in \S{G_A} \,:\, \S{R}_X \cap \S{B}(Z, r) \neq \emptyset}.
\]
Figure \ref{fig:cone} illustrates the cone circumscribing a ball.

\begin{figure}[t]
\centering
\includegraphics[width=\textwidth]{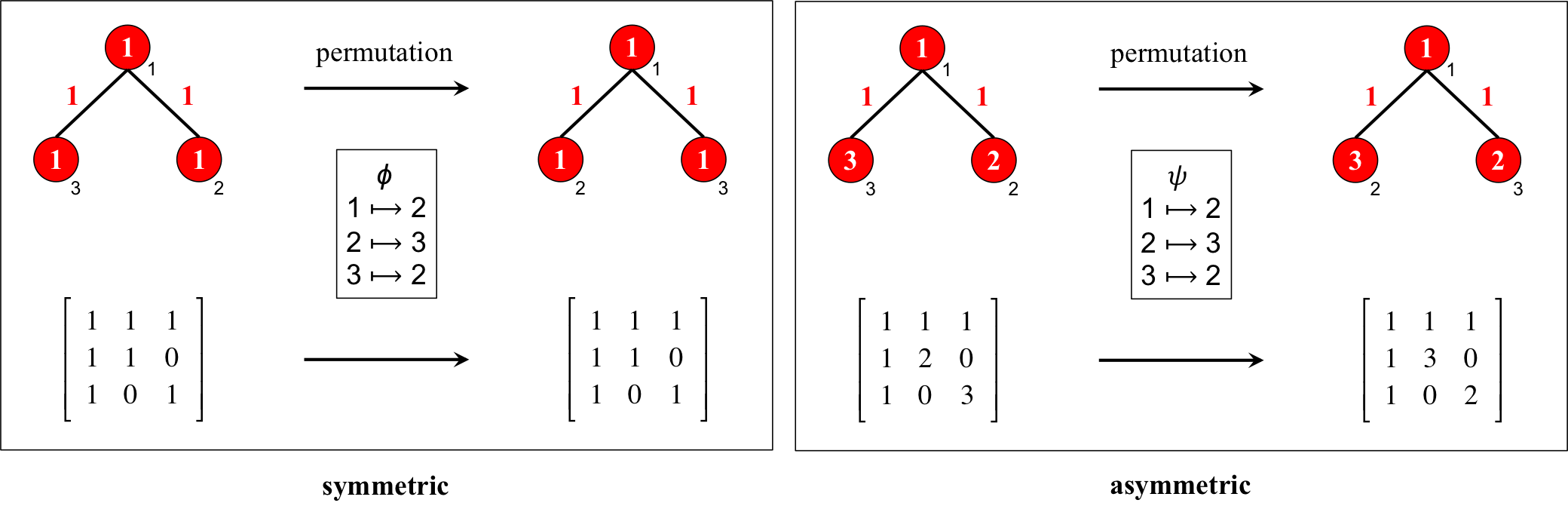}
\caption{\small Left box shows a symmetric and right box an asymmetric graph together with their representation matrices. White numbers inside red balls show the node attributes and red numbers attached to the lines show the edge attributes. Small black numbers attached to each node are its unique identifiers showing the position of the respective node in the representation matrix under consideration. \emph{Left box (symmetric graph):} Permutation $\phi$ in the left box is an example of a non-trivial automorphism that leaves the representation matrix unchanged. \emph{Right box (asymmetric graph):} The node attributes of the asymmetric graph in the right box are mutually different. Any permutation reorders the nodes and results in a different representation matrix. Hence, there is no non-trivial permutation leaving the representation matrix unchanged. The permutation $\psi$ results in a different matrix representation closest to the first one. The degree of asymmetry is $\sqrt{2}$. } 
\label{fig:automorphism}
\end{figure}

\paragraph*{Symmetry, Asymmetrie, and the Degree of Asymmetry.} Let $X = \args{\S{V}, \S{E}, \alpha}$ be a graph. The automorphism group of $X$ consists of all permutations $\phi: \S{V} \rightarrow \S{V}$ such that 
\[
\alpha(i,j) = \alpha(\phi(i), \phi(j)) 
\] 
for all $i,j \in \S{V}$. The automorphism group of $X$ is trivial, if it consists of the identity permutation only. An \emph{asymmetric graph} is a graph with trivial automorphism group. A graph which is not asymmetric is called symmetric. An example of asymmetric graphs are graphs with nodes that have mutually distinct attributes. Figure \ref{fig:automorphism} depicts a a symmetric and an asymmetric graph.

We want to measure how far an asymmetric graph $X$ is from being symmetric. For this, we generalize the degree of asymmetry proposed by Erd\"os and R\'enyi \cite{Erdoes1963} for simple undirected graphs with node and edge attributes from $\cbrace{0,1}$. The generalized \emph{degree of asymmetry} of an asymmetric graph $X$ is given by the quantity 
\[
\chi(X) = \min \cbrace{\norm{\vec{X}^{\phi} - \vec{X}} \,:\, \vec{X}, \vec{X}' \in X \text{ and } \vec{X} \neq \vec{X}' }.
\]
Thus, $\chi(X)$ is the shortest Euclidean distance between two different matrix representations of $X$. In addition, $\chi(X) > 0$ for all asymmetric graphs $X$. The closer an asymmetric graph is to being symmetric, the lower the degree of asymmetry. Therefore, we define $\chi(X) = 0$ for symmetric graphs. 

\paragraph*{Uniqueness of Mean.}
The support of a probability distribution $Q$ on $\S{G_A}$ is the set given by
\[
\S{S}_Q = \cbrace{X \in \S{G_A} \,:\, Q(X) > 0}.
\]
The next result presents sufficient conditions for a unique mean of $Q$.
\begin{theorem}\label{theorem:uniqueness-of-mean}
Let $Q$ be a probability distribution on a graph edit kernel space $\args{\S{G_A}, \delta}$ with support $\S{S}_Q$. The population mean of $Q$ corresponding to a squared loss $L(a)=a^2$ is unique if there is a graph $Z \in \S{G_A}$ with 
\[
\S{S}_Q \subseteq \S{C}_Z = \S{C}\!\args{\!Z,\, \frac{1}{2}\chi(Z)\!}.
\]
\end{theorem}

\noindent
\proof Follows from \cite{Jain2015}, Theorem 3.18. \qed

\bigskip

If $Z$ in Theorem \ref{theorem:uniqueness-of-mean} is symmetric, then the degree of asymmetry is zero. Hence, the cone $\S{C}_Z$ collapses to a ray $\S{R}_Z$ emanating from the zero graph and passing through $Z$. Since it can shown that a ray is isometric to a half line in the Euclidean space, the mean of $Q$ is unique. Otherwise, if $Z$ is asymmetric, the support is contained in a cone $\S{C}_Z$ circumscribing the ball $\S{B}(Z, \chi(Z)/2)$. The larger the degree of asymmetry of $Z$ the wider the cone $\S{C}_Z$. 

Since uniqueness of mean is crucial for comparing distributions \cite{Bhattacharya2013}, we are interested in cones $\S{C}_Z$ with large width and would like to avoid cones that collapse to a single ray passing through a symmetric graph. The next statement says that the subset of symmetric graphs is contained in a null-set and therefore almost all cones $\S{C}_Z$ are non-degenerate. 
\begin{proposition}\label{prop:almost-all-are-asymmetric}
Let $\args{\S{G_A}, \delta}$ be a graph edit kernel space. Then almost all graphs from $\S{G_A}$ are asymmetric. 
\end{proposition}

\noindent
\proof Follows from \cite{Jain2015}, Corollary 3.15. \qed

\bigskip
\noindent
The last result refers to sufficient conditions for uniqueness of sample means. 

\begin{theorem}\label{theorem:uniqueness-of-sample-mean}
Let $\args{\S{G_A}, \delta}$ be a graph edit kernel space and let $\S{S}_n = \args{X_1, \ldots, X_n}$ be a sample of $n$ graphs $X_i \in \S{G_A}$. Then the following statements hold for the sample mean of $\S{S}_n$ corresponding to the squared loss $L(a)=a^2$:
\begin{enumerate}
\item The sample mean is unique for almost all samples $\S{S}_2$. 
\item The sample mean unique if there is a graph $Z$ such that 
\[
\S{S}_n \subseteq \S{C}_Z = \S{C}\!\args{\!Z,\, \frac{1}{2}\chi(Z)\!}.
\]
\end{enumerate}
\end{theorem}

\noindent
\proof Section \ref{proof:theorem:uniqueness-of-sample-mean}. \qed

\subsection{Existence of Midpoints}
Suppose that $X$ and $Y$ are two graphs from $\S{G_A}$. A \emph{midpoint} of graphs $X$ and $Y$ is a graph $M$ satisfying
\[
\delta(X, M) = \delta(Y, M) = \frac{1}{2}\delta(X, Y).
\]
The sample mean and midpoint of two elements coincide in Euclidean spaces and more generally in certain metric spaces, called geodesic spaces. Moreover, existence of a midpoint for every pair of elements is an equivalent characterization of geodesic spaces. Therefore, midpoints in non-metric distance spaces and in graph spaces endowed with the graph edit distance may not exist. 

The next result shows existence of a midpoint and that the notion of midpoint and sample mean of two graphs coincides. 

\begin{corollary}\label{corollary:midpoint}
Let $\args{\S{G_A}, \delta}$ be a graph edit kernel space. For all graphs $X, Y \in \S{G_A}$ the following properties hold:
\begin{enumerate}
\item
$X$ and $Y$ have a midpoint.
\item 
Every midpoint of $X$ and $Y$ is a sample mean. 
\item 
Every sample mean of $X$ and $Y$ is a midpoint.
\end{enumerate}
\end{corollary}

\noindent
\proof Section \ref{proof:corollary:midpoint}. \qed

\subsection{Majorize-Minimize-Mean Algorithm}

As a consequence of Theorem \ref{theorem:nesuco}, we propose a mean algorithm that belongs to the class of Majorize-Minimize / Minorize-Maximize algorithms \cite{Hunter2004}. This class includes the EM algorithm as special case and provides access to general convergence results.

\begin{algorithm}
\caption{\small Majorize-Minimize-Mean (MMM) Algorithm}
\label{alg:mm}
\begin{algorithmic}
\small
\Require
\Statex $\S{S}_n = \args{X_1, \ldots, X_n}$
\Statex \vspace{-0.5em}
\Procedure{}{}
\State choose initial solution $M \in \S{G_A}$ 
\State select representation $\vec{M} \in M$
\Repeat
\ForAll{$i \in \cbrace{1, \ldots, n}$}
\State find $\vec{X}_{\!i} \in X_i$ with $\norm{\vec{X}_{\!i} - \vec{M}} = \delta(X_i, M)$
\EndFor
\Statex \vspace{-1em}
\State set $\displaystyle\vec{M} = \frac{1}{n} \sum_{i=1}^n \vec{X}_i$
\Statex
\Until{convergence}
\EndProcedure
\Statex \vspace{-0.5em}
\Ensure
\Statex representation $\vec{M}$ 
\end{algorithmic}
\end{algorithm}

Algorithm \ref{alg:mm} outlines the basic procedure of the MMM-Algorithm. The algorithm iteratively cycles through the sample $\S{S}_n$ and updates the current solution $M$ according to Theorem \ref{theorem:nesuco}. The basic procedure of the MMM-Algorithm consists of two steps referring to the first two \emph{Ms}. 

\medskip

\noindent
\emph{Step 1 - Majorization:} 
We majorize the sample Fr\'echet function $F_n(Z)$ by a function of the form
\[
f_n(Z) = \sum_{i=1}^n \normS{\vec{X}_{\!i} - \vec{Z}}{^2},
\]
where $\vec{Z}$ is an arbitrarily chosen but fixed matrix representation of $Z$ and the $\vec{X}_i \in X_i$ are representations optimally aligned to representation $\vec{M}$ of the current solution $M$. From definition of the sample Fr\'{e}chet function and the graph edit kernel metric follows
\[
F_n(Z) \leq f_n(Z),
\]
where equality holds at $M$ using representation $\vec{M}$.

\medskip

\noindent
\emph{Step 2 - Minimization:} 
We minimize the majorizing function $f_n(Z)$. The function $f_n(Z)$ is convex and differentiable at $\vec{Z}$ with unique optimal solution 
\[
\vec{M}' = \frac{1}{n} \sum_{i=1}^n \vec{X}_i.
\]
The graph $M'$ represented by $\vec{M}'$ has the property
\[
F_n(M') \leq f_n(M') \leq f_n(M) = F_n(M)
\]
showing that a single iteration of the MMM-Algorithm does not lead to a worse solution. The resulting graph $M'$ with representation $\vec{M}'$ is the current solution of the next iteration. 

The next result establishes convergence of the MMM-Algorithm to a solution satisfying necessary conditions of optimality. In doing so, we eliminate the last remaining anomaly  (P4). The proof adopts Zangwill's Convergence Theorem \cite{Zangwill1969}. 

\begin{theorem}\label{theorem:convergence}
Let $\args{M_k}_{k \geq 0}$ be a sequence of graphs generated by the MMM-Algorithm. 
\begin{enumerate}
\item The limit of any convergent subsequence of $\args{M_k}$ satisfies necessary optimal conditions. 
\item The sequence $\args{F_n(M_k)}$ of sample variations converges.
\end{enumerate}
\end{theorem}
\noindent
\proof Section \ref{proof:theorem:convergence}. \qed

\section{Experiments}\label{sec:Experiments}

The goal of this empircal comparative study is to assess the performance and behavior of the Majorize-Minimize-Mean (MMM) Algorithm.

\subsection{Algorithms}\label{subsec:experiments:algorithms}
We compared the performance of the proposed MMM-Algorithm against the following algorithms: 
\begin{enumerate}
\item Stochastic-Generalized-Gradient Algorithm (SGG)
\item Batch-Arithmetic-Mean (BAM)
\item Incremental-Arithmetic-Mean (IAM)
\item Greedy-Neighbor-Joining (GNJ)
\item Progressive-Alignment-Construction (PAC)
\item \emph{Baseline}: Medoid Algorithm (MED)
\end{enumerate}
Algorithms 1--6 are described in Section \ref{app:sec:algorithms}. The Medoid-Algorithm (MED) serves as baseline for two reasons: (i) it is a very general algorithm that can be applied in any distance space, because it requires no fusion of elements, and (ii) it minimizes the sample Fr\'echet function over the sample itself instead of the whole graph space. As a consequence of the latter issue, MED can not attain a sample mean in general. 

In all experiments, we terminated MMM and SGG after ten iterations waiting time without improvement. For SGG we used a constant step size $\eta$ optimized for each trial over the set $\text{SSG}_\eta = \cbrace{0.9, 0.3, 0.1, 0.07, 0.03, 0.01, 0.007, 0.003, 0.001}$.

\subsection{Data}

\begin{table}[t]
\footnotesize
\centering
\begin{tabular}{ll@{\quad}ccccccc}
\hline
\hline
No & Data $\S{D}$ & Ref. &$\abs{\S{D}}$ & \# classes & $\varnothing \abs{\S{V}}$ & $\max \abs{\S{V}}$ & $\varnothing \abs{\S{E}}$ & $\max\abs{\S{E}}$\\ 
\hline
\\[-2ex] 
1 & Letter (low) & \cite{Riesen2008} & 2,250 & 15 & 4.7 & 8 & 3.1 & 6\\
2 & Letter (medium)& \cite{Riesen2008} & 2,250 & 15 & 4.7 & 9 & 3.2 & 7\\
3 & Letter (high) & \cite{Riesen2008} & 2,250 & 15 & 4.7 & 9 & 4.5 & 9\\
4 & GREC & \cite{Riesen2008} & 1,100 & 22 & 11.5 & 25 & 12.2 & 30\\
5 & AIDS & \cite{Riesen2008} & 2,000 & 2 & 15.7 & 95 & 16.2 & 103\\
6 & MAO	 & \cite{Gauzere2012} & 68 & 2 & 18.4 & 27 & 19.6 & 29\\
7 & Mutag-42 & \cite{Debnath1991} & 42 & 2 & 23.8 & 39 & 25.4 & 43\\
8 & Mutag-188 & \cite{Debnath1991} & 188 & 2 & 26.0 & 40 & 27.9 & 44\\
\hline
\hline
\end{tabular}
\caption{Summary of characteristic features of the graph datasets.}
\label{tab:datasets}
\end{table}

We used the eight datasets shown in Table \ref{tab:datasets} along with some characteristic properties. In the following, we describe the datasets in detail. The description is taken from 
\cite{Riesen2008}, \cite{Gauzere2012}, and \cite{Debnath1991}, respectively. 

\paragraph*{Letter.}
The letter dataset consists of graphs that represent distorted letter drawings of the 15 capital letters A, E, F, H, I, K, L, M, N, T, V, W, X, Y, Z. For each class, a prototype letter was drawn. These prototype letters were converted into prototype graphs by representing lines by undirected edges and ending points of lines by nodes. Each node is labeled with a two-dimensional attribute giving its position relative to a reference coordinate system. Edges are unlabeled. For a given noise level (\emph{low}, \emph{medium}, \emph{high}), $150$ distorted copies of each prototype graph were generated giving a total of $2250$ graphs. The dataset is divided into a training, validation, and test set consisting of $750$ graphs each.

\paragraph*{GREC.}
The GREC dataset consists of graphs representing symbols from architectural and electronic drawings from $22$ classes. The images occur at five different distortion levels. Depending on the distortion level, either erosion, dilation, or other morphological operations were applied. The result was thinned to obtain lines of one pixel width. Finally, graphs were extracted from the resulting denoised images by tracing the lines from end to end and detecting intersections as well as corners. Nodes represent ending points, corners, intersections and circles. Node are labeled with one of the four types of line-segments they represent and with their two-dimensional position. Nodes are connected by undirected edges, which are labeled as line or arc. An additional attribute specifies the angle with respect to the horizontal direction or the diameter in case of arcs. All graphs were distorted nine times to obtain a data set containing $1,100$ graphs uniformly distributed over the $22$ classes. The resulting set is split into a training, validation, and test set of size $286$, $286$, and $528$, respectively.

\paragraph*{AIDS.}
The AIDS data set consists of $2,000$ graphs representing molecular compounds from the AIDS Antiviral Screen Database of Active Compounds. Molecules are divided into two classes: $1,600$ molecules are active inactive against HIV and $400$ are not. The dataset is split into a training, validation, and test set of size $250$, $250$, and $1,500$, respectively.

\paragraph{MAO.}
The Monoamine Oxydase (MAO) dataset consists of $68$ graphs representing molecules divided into two classes: $38$ molecules inhibit the monoamine oxidase (antidepressant drugs) and $30$ do not.

\paragraph{Mutag.}
The mutagenesis dataset consists of $230$ graph representing mutagenic aromatic and heteroaromatic nitro compounds labeled according to whether or not they have a mutagenic effect on the Salmonella typhimurium. The dataset is divided into a subset of $188$ \emph{regression friendly} and a subset of $42$ \emph{regression unfriendly} molecules.

\subsection{Performance Evaluation}

The goal of this series of experiments is to assess the performance and behavior of the proposed MMM-Algorithm with respect to solution quality and runtime .

\subsubsection{Experimental Setup.}

\noindent
\emph{Data}: We sampled graphs from the eight datasets in two different ways: 
\begin{enumerate}
\item \emph{Random Samples}: We randomly sampled $100$ subsets of different size from each of the eight datasets giving a total of $8,000$ samples. The sample sizes were randomly picked from the uniform distribution on $[2, 500]$.
\item \emph{Class Samples}: The samples coincide with the classes of the eight datasets giving a total of $75$ different samples. We replicated each sample $100$ times giving a total of $7,500$ samples. For datasets 1 -- 5, we considered only graphs from the training set. 
\end{enumerate}

\noindent
\emph{Performance Measure}: We used the sample dispersion $\sigma_n = \sqrt{V_n}$ as a measure for solution quality. For runtimes, we used the number of iterations and number of solved graph matching problems as performance measures. For MMM and SGG we recorded the runtime until the best solution was found, that is we did not account for the last ten iterations of waiting time before termination.

\medskip

\noindent
\emph{Parameters}: For SGG, we handled parameter selection of the step size parameter $\eta \in \text{SGG}_{\eta}$ as follows:
\begin{enumerate}
\item \emph{Random Samples}: 
For each sample, we considered the result of SGG with the best sample dispersion over all step sizes $\eta \in \text{SGG}_{\eta}$. 
\item \emph{Class Samples}: 
For each class, we selected the step size that gave the best average sample dispersion over $10$ trials.
\end{enumerate}

\subsubsection{Results and Discussion.}

We present and discuss results on solution quality and runtime.


\paragraph*{Solution Quality} \ 

\begin{figure}[t]
\begin{tabular}{|c|c|}
\hline
&\\[-2ex]
\small Random Samples & \small Class Samples\\[0.5ex]
\hline
&\\
\includegraphics[width=0.49\textwidth]{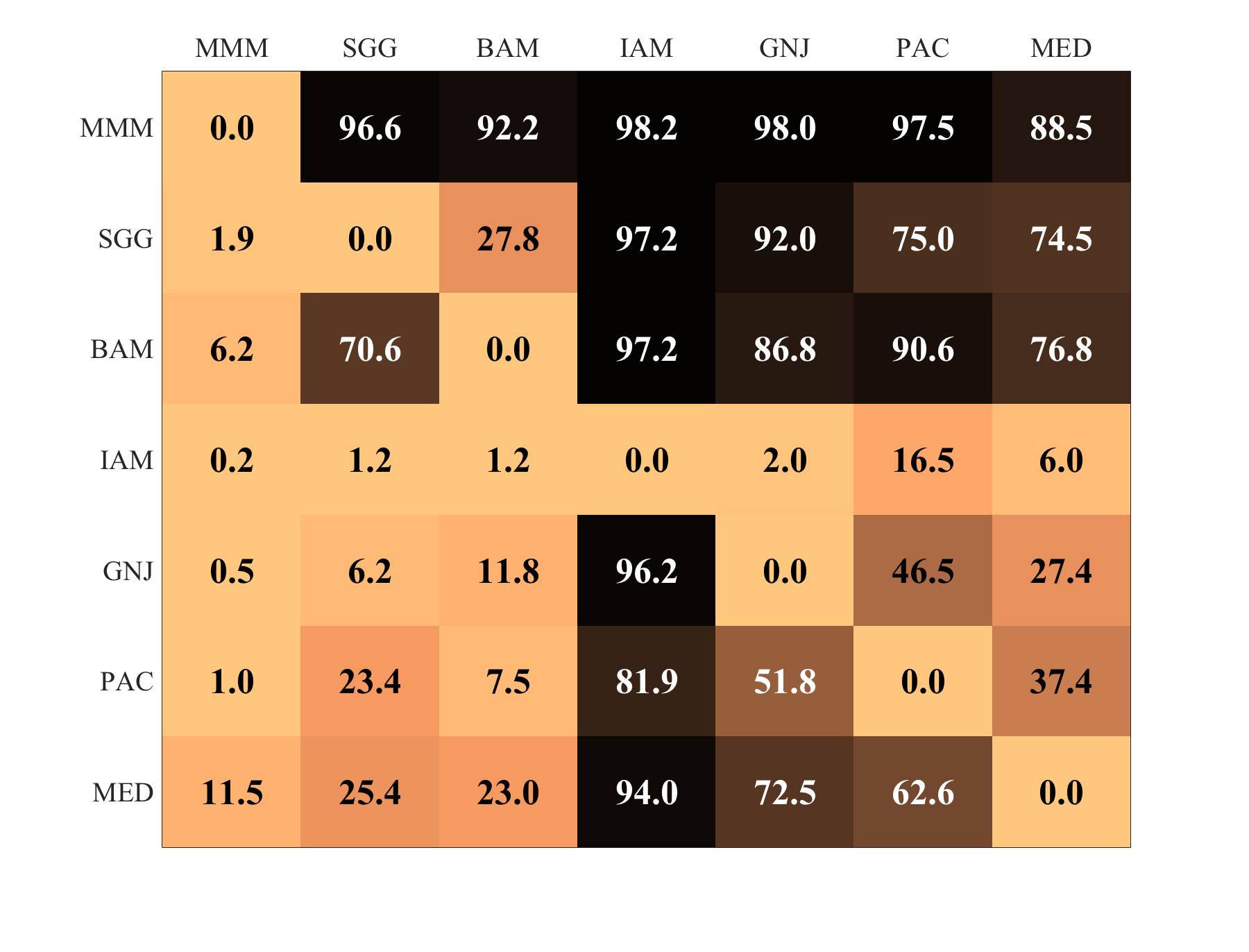} &
\includegraphics[width=0.49\textwidth]{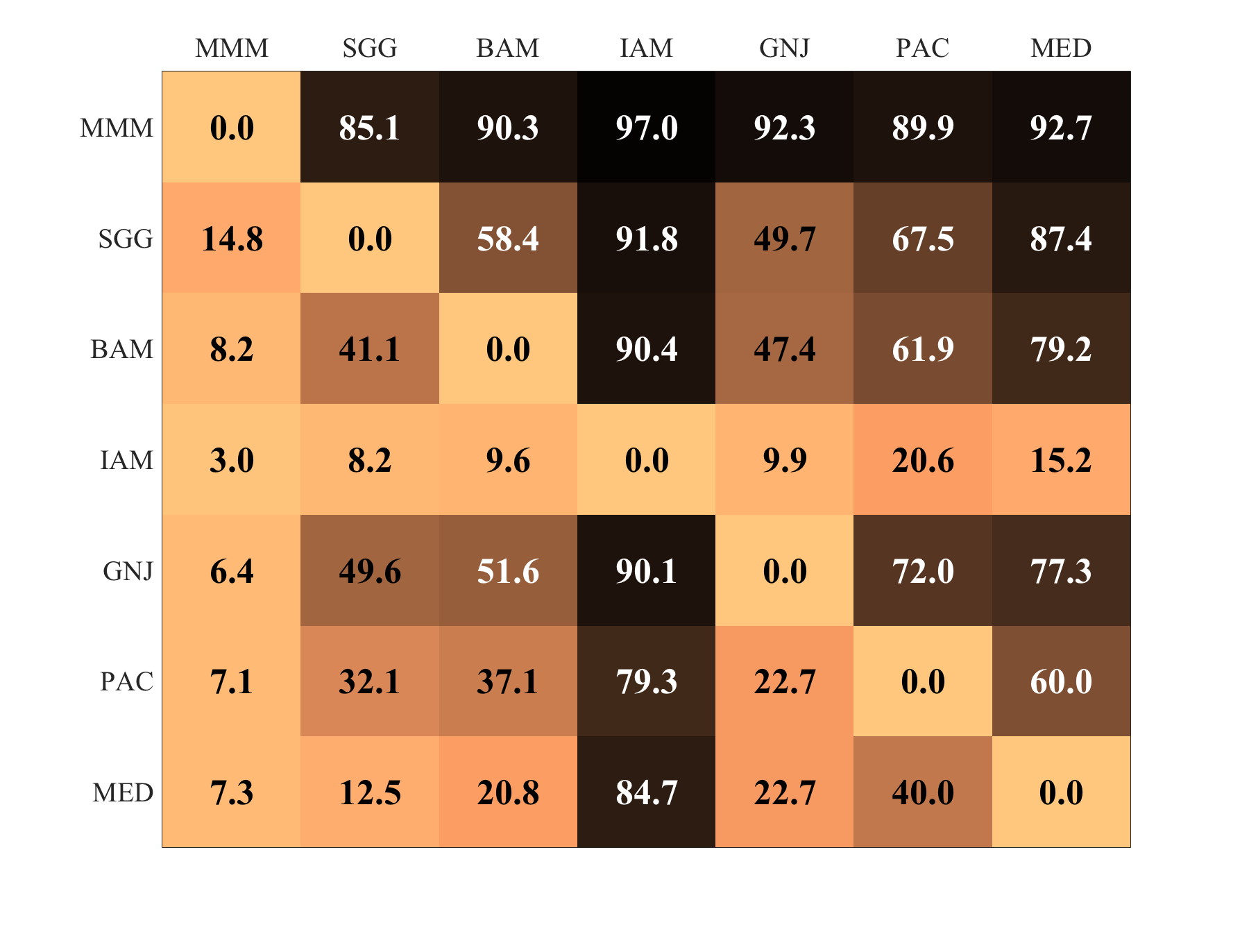} \\
\footnotesize
\begin{tabular}{clcr}
\hline
\hline
\# & ALG & W & \multicolumn{1}{c}{Total} \\
\hline
\\[-2ex]
1. &MMM & 6 & 95.2 \% \\
2. & BAM & 5 & 71.4 \%\\
3. & SGG & 4 &  61.4 \%\\
4. & MED & 3 & 48.2 \%\\
5. & PAC & 2 & 33.8 \%\\
6. & GNJ & 1 & 31.4 \%\\
7. & IAM & 0 & 4.5 \%\\
\hline
\end{tabular}
&
\footnotesize
\begin{tabular}{clcc}
\hline
\hline
\# & ALG & W & \multicolumn{1}{c}{Total} \\
\hline
\\[-2ex]
1. & MMM & 6 & 91.1 \%\\
2. & SGG & 4 & 61.6  \%\\
3. & GNJ & 4 & 57.8 \%\\
4. & BAM & 3 & 54.7 \%\\
5. & PAC & 2 & 39.7 \%\\
6. & MED & 1 & 31.3\%\\
7. & IAM & 0 & 11.1 \%\\
\hline
\end{tabular}
\\
&\\
\hline
\end{tabular}
\caption{Results on pairwise comparisons of seven mean algorithms. \emph{Top row}: Each element $p_{ij}$ of the heatmaps shows the percentage that method in row $i$ has lower (better) sample dispersion than method in column $j$. Percentage of identical sample dispersion for two methods $i$ and $j$ is given by $100-p_{ij} -p_{ji}$. \emph{Bottom row}: The tables show the number of competitions won (W) and the total percentage of pairwise comparisons won for each mean algorithm.}
\label{fig:r_rnk}
\end{figure}

\medskip 

\noindent
Figure \ref{fig:r_rnk} shows the result of pairwise comparisons between the seven mean algorithms using sample dispersion as performance measure.  By inspecting Figure \ref{fig:r_rnk} we made the following observations: 

\begin{enumerate}
\item The results show that MMM won all competitions against the other mean algorithms by a large margin of over 90 \% of all pairwise comparisons. This indicates that MMM will most likely return a better approximation of a sample mean than the other six mean algorithms. Therefore, we suggest MMM as the first choice of a solver when its runtime is not an issue.

\item We observed that Batch-Arithmetic-Mean (BAM) won up to  $8 \%$ of all pairwise comparisons against MMM. At first glance, this result appears surprising, because BAM is a single-loop version of the iterative algorithm MMM. A closer look at the algorithms shows that BAM and MMM randomly select a sample graph as initial solution. This indicates that the choice of initial solution can substantially influence the quality of the final solution in the sense that further iterations can not always compensate a poor initialization. Therefore, we assume that there is room for further improvement of MMM and BAM by considering different initialization schemes.

\item We found that Incremental-Arithmetic-Mean (IAM) lost all six competitions by a similar large margin of about 90 \% of all pairwise comparisons. In addition, IAM performed substantially worse than the other single-loop algorithms BAM, GNJ, and PAC. This indicates that the order by which the sample graphs are fused matters and that a random order is a poor choice.
\end{enumerate}

\begin{figure}[t]
\begin{tabular}{|c|c|}
\hline
&\\[-2ex]
\small Random Samples & \small Class Samples\\[0.5ex]
\hline
\includegraphics[width=0.49\textwidth]{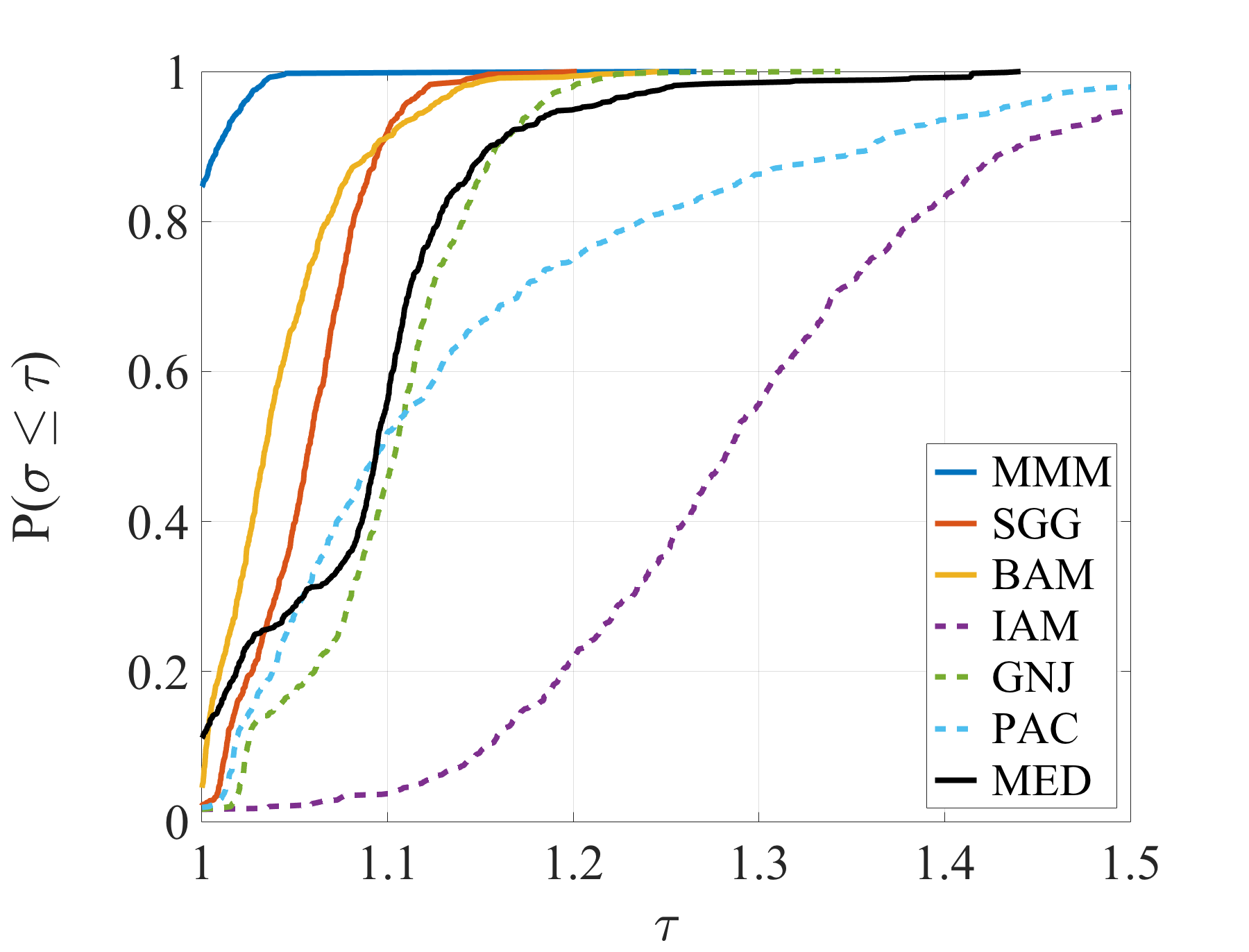} &
\includegraphics[width=0.49\textwidth]{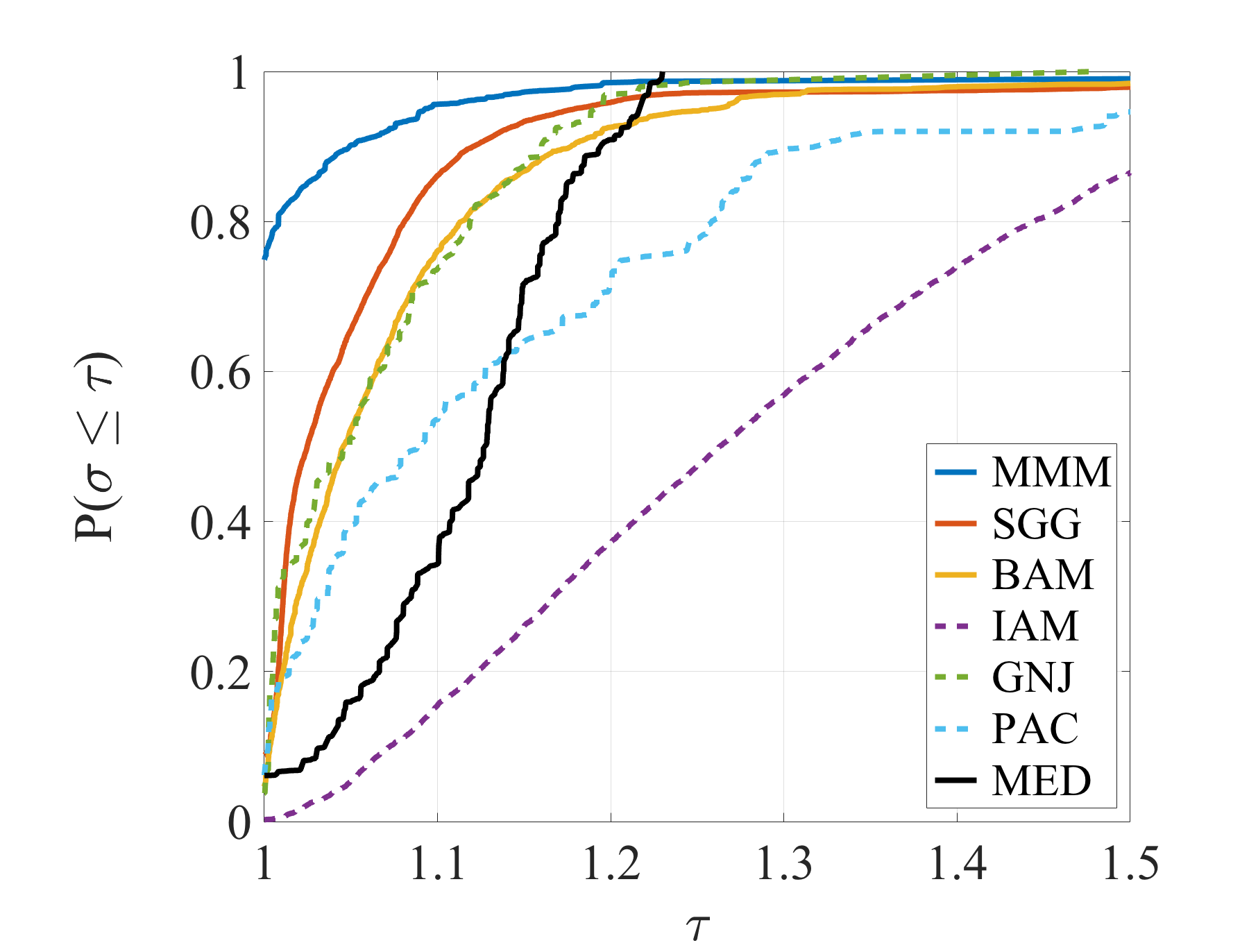} 
\\[1ex]
\footnotesize
\begin{tabular}{clr@{\qquad}clr}
\hline
\hline
\# & ALG & \multicolumn{1}{l}{Wins} & \# & ALG & $\tau_{\max}$ \\
\hline
\\[-2ex]
1. & MMM & 84.6 \% & 1. & SGG & 1.20\\
2. & MED & 11.1 \%  &  2. & BAM & 1.25\\
3. & BAM & 4.5 \%   &  3. & MMM & 1.27\\ 
4. & SGG & 2.1 \%   &  4. & GNJ & 1.34\\ 
5. & PAC & 1.9 \%   &  5. & MED & 1.44\\
6. & GNJ & 1.6 \%  &   6. & IAM & 1.98 \\
6. & IAM & 1.6 \% &    7. & PAC & 2.02\\
\hline
\end{tabular}
&
\footnotesize
\begin{tabular}{clr@{\qquad}clr}
\hline
\hline
\# & ALG & \multicolumn{1}{l}{Wins} & \# & ALG & $\tau_{\max}$ \\
\hline
\\[-2ex]
1. & MMM & 75.0 \% & 1. & MED & 1.23\\
2. & SGG &  8.9 \%  &  2. & GNJ & 1.48\\
3. & PAC & 6.2 \%    &  3. & MMM & 1.61\\ 
4. & MED & 6.1 \%   &  4. & BAM & 1.67\\ 
5. & BAM & 4.7 \%   &  5. & PAC & 1.75\\
6. & GNJ & 3.8 \%    &  6. & SGG & 2.12 \\
7. & IAM & 0.2 \%     &  7. & IMC & 2.89\\
\hline
\end{tabular}
\\
&\\
\hline
\end{tabular}
\caption{Plots in top row show performance profiles using sample dispersion as performance measure. Tables in bottom row show percentage of wins and maximum (worst) factor $\tau_{\max}$. }
\label{fig:r_pp_std}
\end{figure}

Figure \ref{fig:r_rnk} ranks the seven mean algorithms in a pairwise competition but provides no information about differences in solution quality. More generally, rankings provide no information about the percentage of problems a mean algorithm deviates within a given factor from the best solution. Therefore, we use performance profiles introduced by \cite{Dolan2002} to evaluate and compare the solution quality of mean algorithms. We refer to Section \ref{app:sec:performance-profiles} for a brief description of performance profiles.

Figure \ref{fig:r_pp_std} shows the performance profiles of the seven mean algorithms using sample dispersion as performance measure. We made the following observations:

\begin{enumerate}
\item The MMM-Algorithm had the most wins, that is MMM has the highest (estimated) probability of being the optimal solver of all seven mean algorithms (see tables in Figure \ref{fig:r_pp_std}). The probability that MMM wins was about $0.84$ on random samples and about $0.75$ on class samples.
\item\label{enum:obs:MMM_failures} The sample dispersion of MMM deviated from the best solution not more than by a factor of 
\begin{enumerate}
\item 
$\tau = 1.1$ for about $99.8 \%$ ($95.6 \%$) of all random (class) samples,
\item 
$\tau = 1.2$ for about $99.9 \%$ ($98.5 \%$) of all random (class) samples.
\end{enumerate}
Thus, MMM would be the first choice in a practical setting if we only accept solutions that deviate at most by $10 \%$ ($20 \%$) from the best solution with $95 \%$ ($98 \%$) confidence.
\item We observed that the worst factor of MMM was $\tau_{\max} = 1.27$ on random samples and $\tau_{\max} = 1.61$ on class samples. We see that the worst factors of SGG on random samples ($\tau_{\max} = 1.2$) and of MED on class samples ($\tau_{\max} = 1.23$) are lower than  $\tau_{\max}$ of MMM. This shows that MMM is no longer the first choice if we, for example, only accept solutions that deviate at most by $25 \%$ from the best solution with $100 \%$ confidence. These findings indicate that SGG is more robust on random samples and MED is more robust on class samples than MMM.
\item 
Comparison of results across random and class samples shows that there is no mean algorithm that guarantees solutions within a factor of $\tau = 1.4$ with $100 \%$ confidence. Since $40 \%$ deviation from the best solution can be too large for some application problems, we may compromise perfect confidence by a tighter factor. In doing so, we suggest MMM as the first choice in a practical setting due to our findings in item (1) and (2).
\end{enumerate}
\clearpage
\begin{samepage}
\begin{table}[t]
\centering
\footnotesize
\begin{tabular}{cccc}
\hline
\multicolumn{2}{c}{Random Samples} & \multicolumn{2}{c}{Class Samples}\\
\hline
\\[-2ex]
\footnotesize
\begin{tabular}{lr}
\multicolumn{2}{c}{$\tau = 1.1$}\\
\hline
\\[-2ex]
\em Data & \#\\
\hline
\\[-2ex]
Letter (low) & 1 \\
MAO & 1 \\
&\\&\\&\\&\\&\\
\end{tabular}
&
\footnotesize
\begin{tabular}{lrr}
\multicolumn{2}{c}{$\tau = 1.2$}\\
\hline
\\[-2ex]
\em Data & \#\\
\hline
\\[-2ex]
Letter (low)  & 1 \\
&\\&\\&\\&\\&\\&\\
\end{tabular}
&
\footnotesize
\begin{tabular}{lr}
\multicolumn{2}{c}{$\tau = 1.1$}\\
\hline
\\[-2ex]
\em Data & \#\\
\hline
\\[-2ex]
GREC 3 & 100 \\
GREC 5 & 4 \\
GREC 10 & 71 \\
GREC 11 & 17 \\
GREC 13 & 100 \\
GREC 15 & 30 \\
MAO 2 & 6 \\
\end{tabular}
&
\footnotesize
\begin{tabular}{lrr}
\multicolumn{2}{c}{$\tau = 1.2$}\\
\hline
\\[-2ex]
\em Data & \#\\
\hline
\\[-2ex]
GREC 11 & 13 \\
GREC 13 & 100 \\
&\\&\\&\\&\\&\\
\end{tabular}
\\
\hline
\hline
\end{tabular}
\caption{Datasets and number of trials with sample dispersion of MMM worse than $\tau = 1.1$ and $\tau = 1.2$. For class samples, the datasets are broken by classes.}
\label{tab:MMM_failures}
\end{table}

\begin{figure}[h]
\centering
\begin{tabular}{|c|c|}
\hline
&\\[-2ex]
\small Original GREC & \small Normalized GREC \\[0.5ex]
\hline
&\\
\includegraphics[width=0.49\textwidth]{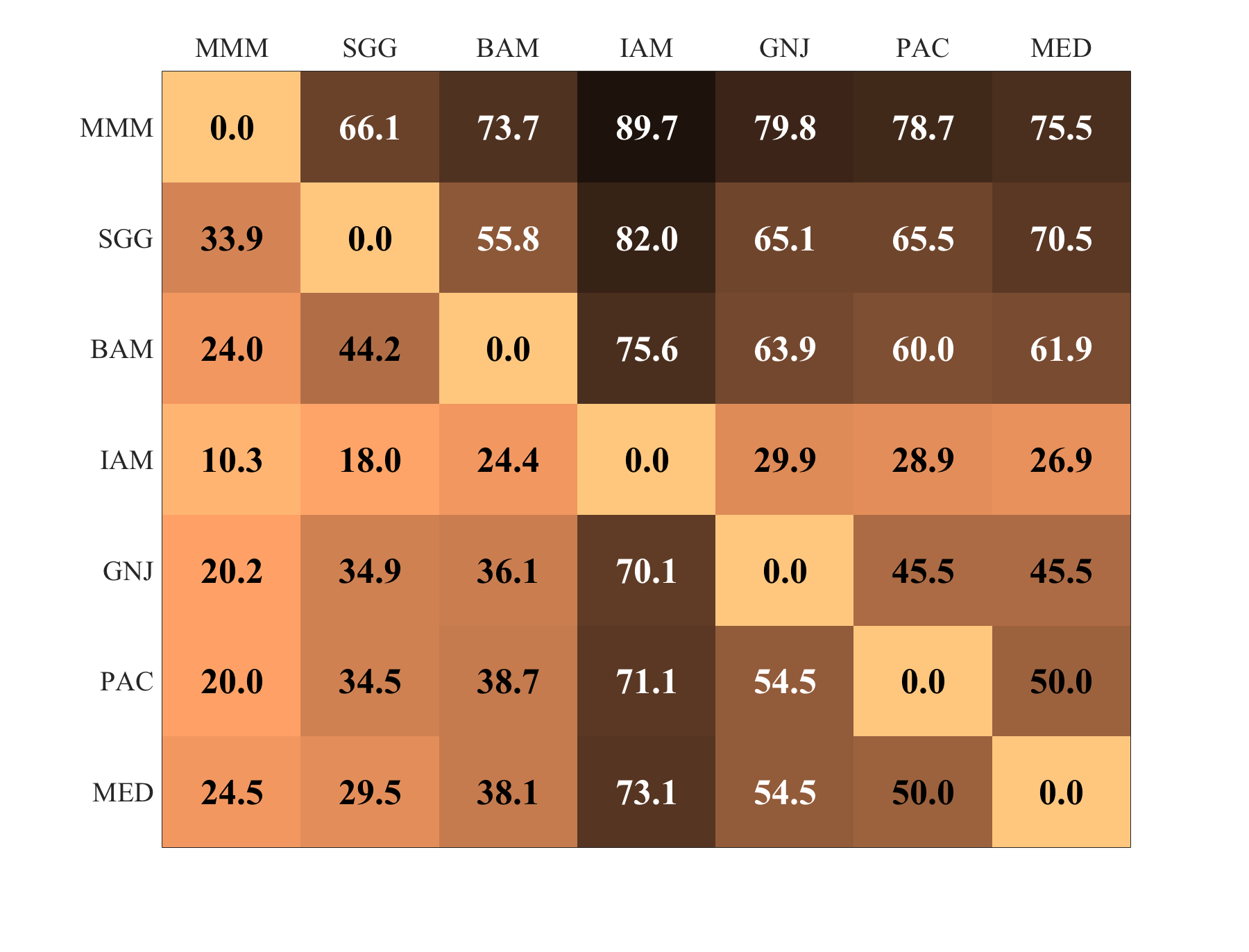} &
\includegraphics[width=0.49\textwidth]{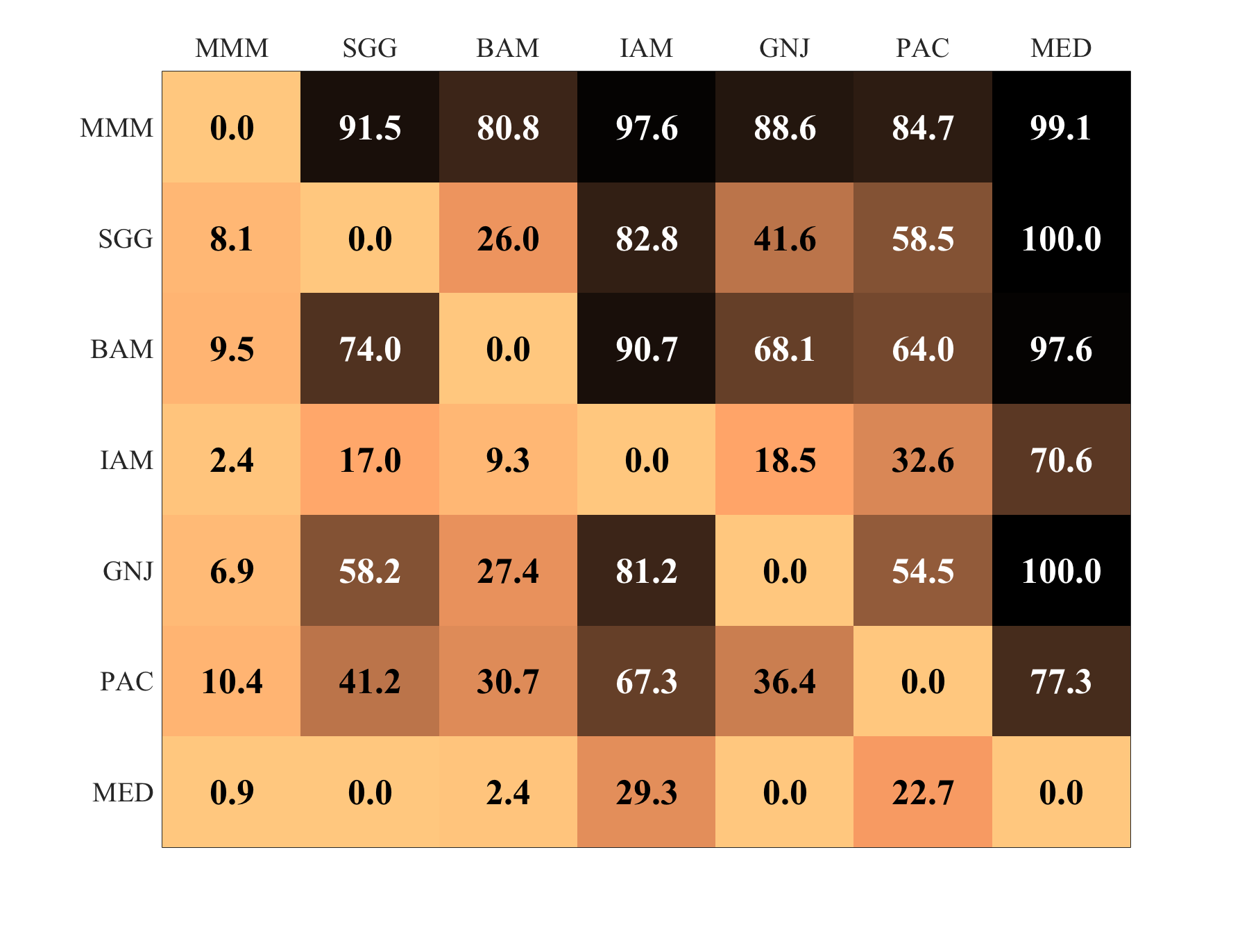} \\
\footnotesize
\begin{tabular}{clcr}
\hline
\hline
\# & ALG & W & \multicolumn{1}{c}{Total} \\
\hline
\\[-2ex]
1. & MMM & 6 & 77.3 \%\\
2. & SGG & 5 & 62.1  \%\\
3. & BAM & 4 & 54.9 \%\\
4. & MED & 2 & 45.0 \%\\
5. & PAC & 2 & 44.8 \%\\
6. & GNJ & 1 & 42.0 \%\\
7. & IAM & 0 & 23.1 \%\\
\hline
\end{tabular}
&
\footnotesize
\begin{tabular}{clcc}
\hline
\hline
\# & ALG & W & \multicolumn{1}{c}{Total} \\
\hline
\\[-2ex]
1. & MMM & 6 & 90.4 \%\\
2. & BAM & 5 & 67.3 \%\\
3. & GNJ & 4 & 54.7 \%\\
4. & SGG & 3 & 52.8  \%\\
5. & PAC & 2 & 43.9 \%\\
6. & IAM & 1 & 25.1 \%\\
7. & MED & 0 & 9.2 \%\\
\hline
\end{tabular}
\\
&\\
\hline
\end{tabular}
\caption{Pairwise comparisons of sample dispersions as described in Figure \ref{fig:r_rnk}. }
\label{fig:r_rnk_grec}
\end{figure}
\end{samepage}

As shown in observation (\ref{enum:obs:MMM_failures}) of Figure \ref{fig:r_pp_std}, MMM failed to solve the problem within factor $\tau = 1.1$ ($\tau = 1.2$) of the best solution in $4.4 \%$ ($1.5 \%$) of all class samples. We are interested in understanding the causes that result in degraded performance of MMM. 

Table \ref{tab:MMM_failures} shows the datasets (and the respective classes) for which MMM failed to find a solution within factor $\tau = 1.1$ and  $\tau = 1.2$, respectively. It is notable that MMM performed worse predominantly on class samples from the GREC dataset. As opposed to other datasets considered in this study, a distinguishing feature of the GREC dataset is a substantially different scaling of the node attributes by up to three orders of magnitude. This is shown by the mean node attribute vector and its standard deviation:
\begin{align*}
\vec{\mu} &= \args{271.6,\;  272.2,\;    0.6,\;   0.3,\;    0.1,\;    0.0}  \\
\vec{\sigma} &= \args{150.6,\;  127.5,\;    0.5,\;    0.4,\;    0.4,\;    0.1 }
\end{align*}

We hypothesize that different scalings of the attributes influences the performance of MMM. To test this hypothesis, we first normalized the node and edge attributes of the GREC graphs to mean zero and standard deviation one for each dimension. Then we replaced the original GREC dataset by the normalized one. Finally, we conducted the same experiment on class samples as before, but restricted to classes from GREC. As control experiment, we repeated the same experiment, where the attributes of the normalized GREC graphs are rescaled by a constant factor $200$. This control experiment serves to exclude the possibility that the effects of degraded performance of MMM were caused by the scale of the attribute values.

Figure \ref{fig:r_rnk_grec} shows the result of pairwise comparisons using sample dispersion as performance measure. The results show that normalization had the strongest positive effect on MMM, BAM, and GNJ and the strongest negative effect on MED and SGG. Rescaling the normalized graph by factor $200$ gave almost the same results as on normalized graphs, where slight differences are caused by the random factors in the experiments. These findings indicate that different scalings of the attribute values can substantially degrade the performance of MMM and other fusion-based mean algorithms.


\paragraph*{Runtime Comparison} \

\medskip

\noindent

\begin{figure}[t]
\centering
\includegraphics[width=0.9\textwidth]{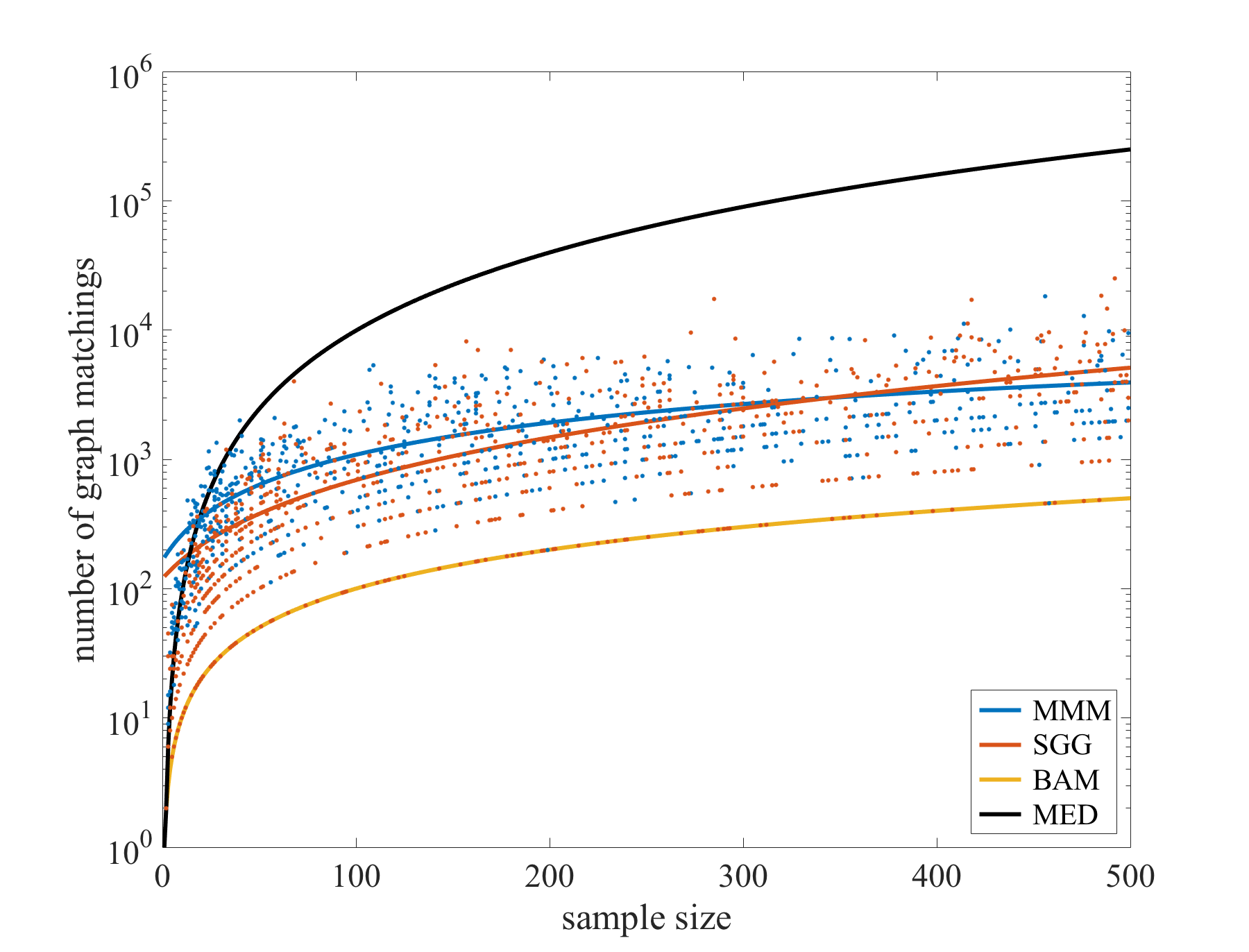} 
\caption{Runtime as a function of sample size. A logarithmic (base 10) scale is used for the vertical axis. The curves for MMM and SGG are polynomials of degree two that best fit the data in a least-squares sense.}
\label{fig:r_xt_by_size}
\end{figure}

\begin{figure}[t]
\centering
\includegraphics[width=0.7\textwidth]{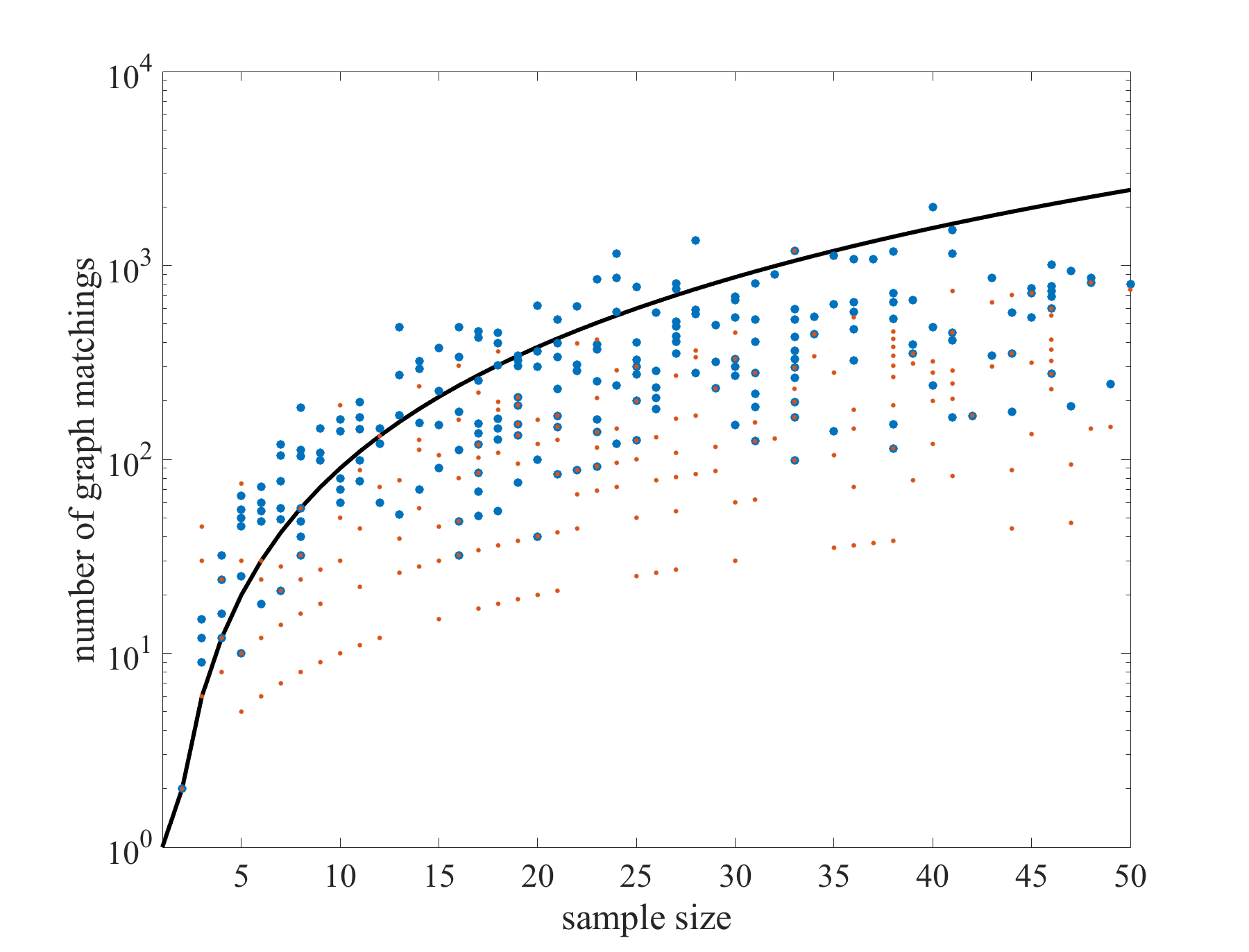} 
\caption{Excerpt of Figure \ref{fig:r_xt_by_size} showing the runtimes as a function of sample sizes. Runtimes of MMM are shown as blue dots, of SGG as red dots, and of MED as black line.}
\label{fig:r_xt_by_smallsize}
\end{figure}

For runtime comparisons, we considered Batch-Arithmetic-Mean (BAM) as a representative of single-loop algorithms with linear time complexity  and the Median Algorithm (MED) as representative of mean algorithms with quadratic time complexity. For details on times complexities, we refer to Section \ref{app:subsec:runtimes}.  

Figure \ref{fig:r_xt_by_size} shows the runtime in logarithmic scale as a function of the sample size. The curves for the MMM-Algorithm and Stochastic-Gradient-Descent (SGG) are obtained by fitting polynomials of degree two. We made the following observations:

\begin{enumerate}
\item We observed that the shape of the quadratic fits of both iterative algorithms, MMM and SGG, is more similar to the linear curve of BAM than to the quadratic curve of MED. This finding indicates that MMM and SGG more likely depend linearly than quadratically on the sample size. 

\item Our results show that MMM is slower on average than SGG on sample sizes up to $300 - 350$ but becomes faster on average for larger sample sizes. This observation suggests that MMM scales better with sample size than SGG. However, only a larger empirical study could determine how both algorithms scale with increasingly large sample size.

\item
The smaller the sample size, the more often MMM was slower than MED. This can be seen more clearly in Figure \ref{fig:r_xt_by_smallsize}. We observed that MMM was slower than MED on samples of average size $13.3$. The average number of iterations of MMM and MED on those samples were $20.1$ and $6.1$, respectively. The results show a similar effect for SGG, but less often. We assume that this effect is caused by a comparably large waiting time of MMM after the best solution was found until termination of MMM.\footnote{Recall that we recorded the runtime of MMM until the best solution was found and used a waiting time of ten iterations.} During this waiting time, the current best solution found so far can be improved, which in turn restarts the waiting time. If the waiting time of $w$ iterations is large compared to the sample size, it may effect the runtime by more than $w$ iterations due to repeated improvements.  A waiting time of ten iterations corresponds to the runtime of MED on samples of size $21$ and can therefore be considered as large for samples of average size $13.3$. For large samples, a waiting time of ten iterations is sufficiently small to prevent quadratic runtimes. This would explain why MMM was not slower than MED for samples of size larger than $40$. 
\end{enumerate}

We conclude with a final remark. The runtime of SGG depends on the choice of step size. The larger the step size, the faster SGG may terminate. To fairly assess the runtime of SGG, we first selected a step size $\eta_*$ from the SGG$_{\eta}$ that gave the best sample variation on average. Then we measured the runtime of SGG with the selected step size $\eta_*$, but ignored the time required to find $\eta_*$.  This issue should be considered when runtime is an issue.

\paragraph*{Summary} \ 

\medskip

\noindent
Key findings are:
\begin{enumerate}
\item MMM found the best approximations of a sample mean than all other mean algorithms in over $75 \%$ of all trials. In a pairwise comparison, MMM was on average better than any other mean algorithm in over $90 \%$ of all trials.
\item There is room for improvement for MMM by studying different initialization schemes.
\item MMM often failed finding good approximations of a sample mean when the scaling of the attributes differed to strongly. If an application admits, normalizing the attribute values in each dimension can help to improve the results. 
\item MMM tends to have linear runtime complexity with respect to the number of graph matchings. 
\end{enumerate}

\medskip

\noindent
Further experiments are necessary to study the following open issues:
\begin{enumerate}
\item How does the waiting time of MMM influence runtime and solution quality?
\item How does the runtime of MMM scale with large sample sizes of more than $500$?
\end{enumerate}

\subsection{Nearest-Neighbor Classification}

The goal of this experiment is to assess the trade-off between classification accuracy and computation time of nearest-neighbor classifiers using condensed datasets for classification. 
We condensed the datasets by reducing the training set to a subset consisting of a sample mean for each class. This empirical study is motivated by the fact that the simple nearest-neighbor classifier still belongs to the state-of-the-art in graph classification, but is computationally too expensive when used with the whole training set for classification.

\subsubsection{Experimental Setup}
We classified the test set of each dataset using eight different nearest-neighbor classifiers. The first nearest-neighbor (NN) classifier used the whole training set for classification of the test set. We refer to this NN classifier as 1-NN. The other seven NN classifiers used a condensed dataset consisting of a sample mean for each class. We used the seven mean algorithms listed in Section \ref{subsec:experiments:algorithms} for approximating a sample mean of each class from the respective training sets. For each dataset and each condensed NN-classifier, we conducted $30$ trials and recorded the average classification accuracy and its standard deviation. 

\subsubsection{Results and Discussion}

\begin{table}[t]
\footnotesize
\centering
\begin{tabular}{l@{\qquad}rrrrrrrr}
\hline
\hline
& 1-NN & MMM & SGG & BAM & IAM & GNJ & PAC & MED \\
\hline
\\[-2ex]
Letter (low) 		& \bf {\color{blue}96.9} & \bf \em{\color{red}96.0}& 95.9& 95.9 & 90.0 & \bf \em{\color{red}96.0} &  94.5 & 95.1\\
				&         & $^{\pm 0.0}$ & $^{\pm 0.2}$& $^{\pm 0.1}$& $^{\pm 4.1}$ & $^{\pm 0.0}$& $^{\pm 0.0}$&  $^{\pm 0.0}$ \\
Letter (medium) 	& 92.4 & 92.4& \bf \bf {\color{blue}93.0}& 92.3 & 82.7 & \bf\em {\color{red}92.7} & 92.9 & 90.3\\
				&         & $^{\pm 0.1}$ & $^{\pm 0.3}$& $^{\pm 0.2}$& $^{\pm 4.4}$ & $^{\pm 0.0}$& $^{\pm 0.0}$&  $^{\pm 0.0}$ \\
Letter (high) 		& \bf\em {\color{red}87.6} & \bf {\color{blue}87.9}& 81.9& 87.1 & 48.0 & 82.0 & 69.6 & 79.3\\
				&          & $^{\pm 0.1}$ & $^{\pm 1.6}$& $^{\pm 0.7}$& $^{\pm 3.7}$ & $^{\pm 0.0}$& $^{\pm 0.0}$&  $^{\pm 0.0}$ \\
GREC 			& \bf {\color{blue}98.7} & 84.6& 85.9& 84.7 & 81.5 & \bf\em{\color{red}86.2} & 85.0 &  83.3\\
				&         & $^{\pm 0.7}$ & $^{\pm 1.2}$& $^{\pm 0.7}$& $^{\pm 2.9}$ & $^{\pm 0.0}$& $^{\pm 0.0}$&  $^{\pm 0.0}$ \\
AIDS 			& \bf {\color{blue}99.5} & 90.9& 90.5& 90.9 & 88.7 & 89.6 & 90.8 & \bf\em{\color{red}91.2}\\
				&         & $^{\pm 0.1}$ & $^{\pm 0.5}$& $^{\pm 0.0}$& $^{\pm 0.7}$ & $^{\pm 0.0}$& $^{\pm 0.0}$&  $^{\pm 0.0}$ \\
MAO	 			& \bf\em{\color{red}86.8} & \bf {\color{blue}88.2}& 76.8& 85.5 & 66.8 & 69.1 & 63.2 & 64.7\\
				&         & $^{\pm 0.0}$ & $^{\pm 3.5}$& $^{\pm 2.6}$& $^{\pm 3.8}$ & $^{\pm 0.0}$& $^{\pm 0.0}$&  $^{\pm 0.0}$\\
Mutag-42 			&  \bf\em{\color{red}76.2} & \bf{\color{blue}80.3}& 76.0&  \bf\em{\color{red}76.2} & 60.8 & 69.0 & 33.3 & 54.8\\
				&         & $^{\pm 3.4}$ & $^{\pm 5.4}$& $^{\pm 3.8}$& $^{\pm 6.9}$ & $^{\pm 0.0}$& $^{\pm 0.0}$&  $^{\pm 0.0}$ \\
Mutag-188 		& \bf {\color{blue}89.6} & \bf\em{\color{red}81.2}& 76.7 & 79.5 & 54.0 & 78.4 & 58.6 & 79.7\\
				&         & $^{\pm 1.5}$ & $^{\pm 2.4}$& $^{\pm 2.0}$& $^{\pm 3.1}$ & $^{\pm 1.6}$& $^{\pm 11.0}$&  $^{\pm 1.4}$ \\
\hline
\\[-2ex]
\em Wins (all) 		& 4 & 3 & 1 & 0 & 0 & 0 & 0 & 0\\   	
\em Wins (condensed) & -- & 5 & 1 & 0 & 0 & 2 & 0 & 1\\  	 
\hline
\hline
\end{tabular}
\caption{Average classification results and standard deviation. Overall best results are highlighted in bold-blue and second best in italics-red.}
\label{tab:r_classification}
\end{table}

Table \ref{tab:r_classification} shows the average classification accuracies and standard deviations. We made the following observations:
\begin{enumerate}
\item Compared with the condensed NN-classifiers, the results show that the MMM-based condensed NN-classifier won on five out of eight datasets. In addition, MMM won on three datasets against the best performing 1-NN.  This indicates that a good approximation of a sample mean more likely results in a good condensed NN-classifiers. This argument is also supported by the poor results of IAM and PAC. Both mean-algorithm performed worst in finding good approximation of a sample mean and both algorithms resulted in condensed 1-NN with the worst average classification accuracy, in particular for the datasets Letter (high), MAO, Mutag-42, and Mutag-188.

\item The results show that condensed NN-classifiers won on half of the datasets against 1-NN ($3 \times$ MMM, $1 \times$ SGG). In these cases, condensed NN-classifiers give comparable or even better results than 1-NN but are substantially faster. For classification, 1-NN compares a test graph with all graphs from the training set, while condensed 1-NN classifiers compare the test graphs with the class means only. 

\item We observed that the losses in accuracy incurred by condensing the classes to a sample mean were larger than $10 \%$ for GREC ($-12.5 \%$) and larger than $5 \%$ for AIDS ($-8.3 \%$) and Mutag-188 ($-8.4 \%$). This finding indicates that summarizing classes to a single prototype can result in an uncontrolled loss of relevant information for nearest neighbor classification. We propose two solutions to this problem: (i) centroid-based clustering using $k$ sample means per class as proposed and theoretically justified in \cite{Jain2011} and (ii) adaption of supervised  learning vector quantization algorithms for graphs as proposed and theoretically justified in \cite{Jain2010b,Jain2010c}.
\end{enumerate}

\section{Conclusion}\label{sec:Conclusion}

In a graph edit kernel space the sample mean exists and is a consistent estimator of the population mean. In addition, for two graphs, the concept of sample mean and midpoint coincide. The necessary conditions of optimality describe the form of a sample mean, give rise to Majorize-Minimize-Mean Algorithm that outperformed other mean algorithms based on fusing graphs via optimal alignments, and provide a unifying scheme of those fusion-based algorithms. 

This contribution serves as a first step towards a theory of statistical analysis of graphs. The ultimate goal is to adapt standard tools from traditional statistics and to develop novel tools for inferring properties of populations in graph edit kernel spaces. Possible next steps include adaption of the normal distribution, the Central Limit Theorem, parameter estimation using the maximum likelihood, and principal component analysis.

\begin{small}
\begin{appendix}

\section{Performance Profiles}\label{app:sec:performance-profiles}

To compare the performance of the mean algorithms, we used performance profiles \cite{Dolan2002}. A performance profile is a cumulative distribution function for a performance metric. We considered two performance metrics: (1) the average sample dispersion $\sigma_n = \sqrt{V_n}$, where $V_n$ is the sample dispersion, and (2) the number $t$ of iterations, where one iteration is a loop through a sample $\S{S}_n$. 

To define a performance profile, we assume that $\mathbb{A}$ is the set of all five mean algorithms and $\mathbb{S}$ is the set of all samples. For each sample $s \in \mathbb{S}$ and each mean algorithm $a \in \mathbb{A}$, we define
\[
p_{a,s} = \text{performance of mean algorithm $a \in \mathbb{A}$ on sample $s \in \mathbb{S}$},
\]
where the performance measure is either the sample dispersion or the number of iterations. The performance ratio of mean algorithm $a$ on sample $s$ is given by
\[
r_{a, s} = \frac{p_{a,s}}{\min \cbrace{ p_{\alpha, s} \,:\, \alpha \in \mathbb{A}}}.
\]
The performance profile of mean algorithm $a \in \mathbb{A}$ over all samples $s \in \mathbb{S}$ is an empirical cumulative distribution function
\[
P_a(\tau) = \frac{1}{\abs{\mathbb{S}}} \abs{\cbrace{s \in \mathbb{S} \,:\, r_{a,s} \leq \tau}}.
\]
Then $P_a(\tau)$ is the estimated probability that a solution from algorithm $a$ will not deviate more than by the factor $\tau$ from the best solution over all algorithms under consideration. The value $P_a(1)$ is the estimated probability that algorithm $a$ will win over all other algorithms.

\section{Mean Algorithms}\label{app:sec:algorithms}

This section describes six mean algorithms based on graph fusion \cite{Jain2009} and provides their runtimes with respect to the number of graph matching problems solved.

\subsection{Algorithms}

\paragraph*{Medoid Algorithm.} The \emph{Medoid Algorithm} (MED) picks the minimum of the sample Fr\'echet function over the sample $\S{S}_n$ instead of the whole space $\S{G_A}$. For this, MED computes a distance matrix $\vec{D} = (\delta_{ij})$ whose elements $\delta_{ij} $ are the pairwise squared distances $\delta^2(X_i, X_j)$ between the sample graphs $X_i$ and $X_j$. Then a medoid of $\S{S}_n$ is a sample graph $X_{b}$ satisfying 
\[
b = \arg\min_{i} \sum_{j=1}^n \delta_{ij}.
\]
Then the graph $M_n = X_b$ serves as approximation of a sample mean of $\S{S}_n$. By definition, the sample Fr\'echet variation at a medoid is an upper bound of the sample 
Fr\'echet variation at a sample mean. Furthermore, MED can be applied in any distance space, because it requires no fusion techniques for constructing new elements from given ones. 
Here, we use MED as baseline.

\paragraph*{Incremental-Arithmetic-Mean.} The \emph{Incremental-Arithmetic-Mean Algorithm} (IAM) emulates the incremental computation of the mean of $n$ numbers $x_1, \ldots, x_n \in \R$, when the numbers $x_i$ are drawn one-by-one.
\begin{align*}
\mu_k &= \begin{cases}
x_1 & k = 1\\[1ex]
\displaystyle \frac{k-1}{k}\,\mu_{k-1} + \frac{1}{k}\,x_k & 1 < k \leq n\\
\end{cases}.
\end{align*}
Then the mean of the $n$ numbers is given by $\mu = \mu_n$. To adopt the incremental mean computation for graphs, we rewrite $\mu_k$ by
\begin{align*}
\mu_k &= \frac{k-1}{k}\,\mu_{k-1} + \frac{1}{k}\,x_k\\
&= \mu_k - \frac{1}{k}\args{x_k - \mu_k}
\end{align*}
The last equation has the form of an incremental update rule with decreasing step size $\eta_k = 1/k$. Thus, we obtain IMC from the SGG-Algorithm by choosing a decreasing step size $\eta_k = 1/k$ and terminating after a single loop through the sample $\S{S}_n$. The solutions of the IAM-Algorithm depend on the order the sample graphs are processed.

\paragraph*{Batch-Arithmetic-Mean.}  The \emph{Batch-Arithmetic-Mean Algorithm} (BAM) emulates the standard mean computation of $n$ numbers $x_1, \ldots, x_n \in \R$ given by
\[
\mu = \frac{1}{n} \sum_{i=1}^n x_i.
\]
We obtain BAM by terminating the proposed MM-Algorithm after a single loop through the sample $\S{S}_n$.

\paragraph*{Greedy-Neighbor-Joining.} The \emph{IGreedy-Neighbor-Joining} (GNJ) Algorithm combines the Medoid Algorithm with the Incremental-Arithmetic-Mean. In doing so, GNJ first determines a medoid $M_n'$ of $\S{S}_n$. To compute a sample mean $M_n$, GNJ invokes Incremental Mean Computation where the sample graphs are presented in the order of increasing distances from the medoid $M_n'$. When used in prototype-based clustering, pairwise distance calculations is dropped and the medoid is replaced by the current centroid of the respective cluster.

\paragraph*{Progressive-Alignment-Construction.} The \emph{Progressive-Alignment-Construction} (PAC) algorithm first computes a pairwise distance matrix as in the Medoid Algorithm. Then the graphs are clustered using agglomerative single linkage clustering \cite{Hastie2009}. Starting with $n$ clusters consisting of singletons, the two closest clusters are merged until all graphs are in a single cluster. During this process, each cluster is represented by a sample mean. When merging two clusters $\S{C}_i$ and $\S{C}_j$ with respective sample means $M_i$ and $M_j$, the mean $M$ of the merged cluster $\S{C} = \S{C}_i \cup \S{C}_j$ is represented by a matrix of the form 
\[
\vec{M} = \frac{\abs{\S{C}_i}}{\abs{\S{C}}}\vec{M}_i + \frac{\abs{\S{C}_j}}{\abs{\S{C}}}\vec{M}_j,
\]
where the representations $\vec{M}_i \in M_i$ and $\vec{M}_j\in M_j$ are optimally aligned to $\vec{M}$.

\paragraph*{Stochastic Generalized Gradient.} 
The  \emph{Stochastic-Generalized-Gradient Algorithm} (SGG) minimizes the non-differentiable sample F\'echet function in a similar fashion as stochastic gradient descent.

The squared graph edit kernel metric $\delta^2$ as a pointwise minimum of convex differentiable functions is non-differentiable and non-convex. But it can be shown that $\delta^2$ is differentiable almost everywhere and locally Lipschitz at non-differentiable points \cite{Jain2009,Jain2011}. Furthermore, $\delta^2$ is generalized differentiable in the sense of Norkin and Ermoliev \cite{Ermoliev1998,Norkin1986}. By the calculus of generalized differentiable function, the (sample) Fr\'{e}chet function is also generalized differentiable. The update rule of the SGG-Algorithm replaces the gradient of the stochastic gradient descent method by an element of the generalized gradient of $\delta^2(X, M)$ given by
\[
\vec{G} = -2\args{\vec{X} - \vec{M}},
\] 
where representation $\vec{X} \in X$ is optimally aligned to $\vec{M} \in M$ \cite{Jain2009,Jain2011}. Then the update rule is of the form
\[
\vec{M} \leftarrow \vec{M} + \eta \vec{G},
\]
where $\eta > 0$ is the step size. At differentiable points the direction $\vec{G}$ coincides with the gradient of $\delta^2(X, M)$ with respect to $M$. At non-differentiable points $-\vec{G}$ need not to be a direction of descent. Therefore, it is common to keep track of the best solution $V_t^*$ found so far after each cycle $t$ through the sample $\S{S}_n$. We set
\[
V_t^* = \min \cbrace{V_{t-1}^*, V_t}
\]
where $V_t = F_n(M_t)$ is the sample variation of the sample Fr\'echet function at the solution after the $t$-th cycle through $\S{S}_n$. In addition we keep record of the best approximation of a sample mean 
\[
M_t^* = \begin{cases}
M_t & V_t < V_{t-1}^*\\
M_{t-1}^* & \text{otherwise}
\end{cases}.
\]
To establish convergence and consistency results, step sizes are square summable but not summable, that is
\[
\sum_{k=1}^\infty \eta_k^2 < \infty, \qquad \sum_{k=1}^\infty \eta_k = \infty,
\]
where $k$ refers to the number of sample graphs that has been randomly picked and presented to the SGG-Algorithm.
 
When the sample graphs are randomly drawn from the underlying probability distribution, the SGG-Algorithm directly minimizes the Fr\'{e}chet function. Then under mild assumptions, the SGG-Algorithm converges almost surely to a solution of the Fr\'{e}chet function satisfying necessary conditions of optimality almost surely \cite{Ermoliev1998,Norkin1986}.

\subsection{Runtime of Mean Algorithms}\label{app:subsec:runtimes}

The main contribution to the runtime of a mean algorithm is the number of graph matching problems that need to be solved to minimize a sample Fr\'echet function. We consider other computations as negligible. Then the runtime of the mean algorithms is as follows:

\begin{center}
\begin{tabular}{llll}
\hline
\hline
Class & Method & Best Case & Worst Case\\
\hline
\\[-2ex]
Iterative & MMM & $\S{O}(n)$ & $\S{O}(Tn)$\\
Iterative &SGG  & $\S{O}(n)$ & $\S{O}(Tn)$\\
Single-Loop & BAM  &  $\S{O}(n)$ & $\S{O}(n)$\\
Single-Loop &IAM  &  $\S{O}(n)$ & $\S{O}(n)$\\
Single-Loop &GNJ & $\S{O}(n)$ & $\S{O}(n^2)$\\
Single-Loop &PAC & $\S{O}(n^2)$ & $\S{O}(n^2)$\\
Baseline& MED & $\S{O}(n^2)$ & $\S{O}(n^2)$\\
\hline
\hline
\end{tabular}
\end{center}
The number $n$ refers to the sample size. The value $T$ is the number of iterations (loops through the whole sample). The complexity of the Greedy-Neighbor-Joining (GNJ) algorithm depends on the choice of reference graph. Here, we consider the medoid as reference graph. Then GNJ requires $n(n-1)/2$ graph matchings for identifying the medoid and another $n-1$ graph matchings for fusing the other sample graphs with the reference graph.  Choosing a random sample graph or the centroid in graph k-means reduces the complexity of GNJ to $\S{O}(n)$.

\section{Proofs}

\subsection{Proof of Theorem \ref{theorem:nesuco}}\label{app:proof-theorem:nesuco}
Let $\S{X}$ be the set of all matrices that represent graphs from $\S{G_A}$. Consider the set 
\[
\S{A}_n(\vec{X}) = \S{A}_1(\vec{X}) \times \cdots \times \S{A}_n(\vec{X}), 
\]
where $\S{A}_i(\vec{X})$ is the set of all representations $\vec{X}_{\!i} \in X_i$ optimally aligned to $\vec{X}$. For the sake of mathematical convenience, we rewrite the sample Fr\'echet function into the equivalent form
\[
F_n: \S{X} \rightarrow \R, \quad \vec{Z} \mapsto \sum_{i=1}^n \min \cbrace{\normS{\vec{X}_{\!i} - \vec{Z}}{^2} \,:\, \vec{X}_{\!i} \in X_i}.
\]
Let $\vec{X} \in \S{X}$ and $\mathbb{X} = \args{\vec{X}_1, \ldots, \vec{X}_n} \in \S{A}_n(\vec{X})$. Then the function
\[
f(\vec{Z}) = f(\vec{Z} ; \vec{X}, \mathbb{X}) = \sum_{i=1}^n \normS{\vec{X}_i - \vec{Z}}{^2}
\]
majorizes the sample Fr\'echet function $F_n(\vec{Z})$ such that $F_n(\vec{Z}) \leq f(\vec{Z})$ for all $\vec{Z} \in \S{X}$, where equality holds at $\vec{X}$, that is $F_n(\vec{X}) = f(\vec{X})$. The function $f(\vec{Z})$ is convex and differentiable with unique minimum 
\[
\vec{X}_* = \frac{1}{n}\sum_{i=1}^n\vec{X}_i,
\]
where $\vec{X}_i \in \S{A}_i(\vec{X})$. Suppose that $\vec{M}$ is a local minimum of $F_n(\vec{Z})$. By the same construction, $f(\vec{Z}) = f(\vec{Z}; \vec{M}, \mathbb{M})$ is a majorizing function of $F_n(\vec{Z})$ with unique minimum
\[
\vec{M}_* = \frac{1}{n}\sum_{i=1}^n\vec{X}_i,
\]
where $\vec{X}_i \in \S{A}_i(\vec{M})$. Then we have $F_n(M) = f(\vec{M}) \leq f(\vec{M}_*)$. 

Suppose that $\vec{M} \neq \vec{M}_*$. Then $\vec{M}$ is not a minimizer of $f(\vec{Z})$. Then for every $\varepsilon > 0$ there is a $\vec{Z} \in \S{B}(\vec{M}, \varepsilon)$ such that 
\[
F_n(\vec{Z}) \leq f(\vec{Z}) < f(\vec{M}).
\]
This contradicts our assumption that $M$ is a local minimum of $F_n$. Thus, we have $\vec{M} = \vec{M}_*$, which completes the proof.

\subsection{Proof of Theorem \ref{theorem:uniqueness-of-sample-mean}}\label{proof:theorem:uniqueness-of-sample-mean}

The second assertion follows from Theorem \ref{theorem:uniqueness-of-mean}. We prove the first assertion. For this, we need the notion of Dirichlet cell. The Dirichlet cell of graph $Z$ centered at matrix $\vec{Z}$ is defined by
\[
\S{D}(\vec{Z}) = \cbrace{\vec{X} \in X \,:\, \delta(X, Z) = \norm{\vec{X} - \vec{Z}} \text{ for all } X \in \S{G_A}}.
\]
If $Z$ is asymmetric, its Dirichlet cells are fundamental domains. 

Let $X, Y$ be asymmetric graphs with matrix representations $\vec{X}$ and $\vec{Y}$ such that $\vec{Y}$ lies in the interior of $\S{D}(\vec{X})$. From Prop.~\ref{prop:almost-all-are-asymmetric} and \cite{Jain2015}, Theorem 3.16 follows that almost all pairs $X, Y$ of graphs satisfy these properties. It remains to show that the sample mean of $X$ and $Y$ is unique. 

First, we show that there is a representation of the sample mean $M$ that lies in the interior of $\S{D}(\vec{X})$. By construction, we have $\delta(X, Y) = \norm{\vec{X} - \vec{Y}}$. Then 
$\vec{M} = (\vec{X} + \vec{Y})/2$ is a sample mean of $\vec{X}$ and $\vec{Y}$. Since $\S{D}(\vec{X})$ is convex by \cite{Jain2015}, Theorem 3.16, the mean $\vec{M}$ lies in the interior of $\S{D}(\vec{X})$. This implies $\delta(X, M) = \norm{\vec{X} - \vec{M}}$, where $M$ is the graph represented by $\vec{M}$. Since $\vec{M}$ is a midpoint of $\vec{X}$ and $\vec{Y}$, we have
\[
\delta(X, M) = \norm{\vec{X} - \vec{M}} = \norm{\vec{Y} - \vec{M}} = \frac{1}{2}\norm{\vec{X}-\vec{Y}} = \frac{1}{2}\delta(X, Y).
\]
Suppose that $\delta(Y, M) < \norm{\vec{Y} - \vec{M}}$. Then we obtain the contradiction
\[
\delta(X, Y) \leq \delta(X, M) + \delta(Y, M) < \norm{\vec{X} - \vec{M}} + \norm{\vec{Y} - \vec{M}} = \norm{\vec{X}-\vec{Y}} = \delta(X,Y). 
\]
Thus, we have $\delta(Y, M) = \norm{\vec{Y} - \vec{M}}$. Then $M$ is a midpoint of $X$ and $Y$. From Corollary \ref{corollary:midpoint} follows that $M$ is a sample mean of $X$ and $Y$.

Suppose that $M'$ is another sample mean of $X$ and $Y$ distinct from $M$. By Theorem \ref{theorem:nesuco} there are representations $\vec{X}' \in X$ and $\vec{Y}' \in Y$ such that 
$\vec{M}' = \args{\vec{X}' + \vec{Y}'}/2$ is a representation of $M'$. Let $\S{G}$ be the group of node permutations that simultaneously permutes row and columns of matrix representations. Then there is a $\gamma \in \S{G}$ such that $\vec{X}' = \gamma \vec{X}$. Since $X$ is asymmetric, we have $\S{D}(\vec{X}') = \gamma \S{D}(\vec{X})$ by \cite{Jain2015}, Prop.~3.13. Since $\vec{Y}$ is in the interior of $\S{D}(\vec{X})$, so is $\gamma \vec{Y}$ in the interior of $\gamma \S{D}(\vec{X})$. From \cite{Jain2015}, Prop.~3.13 follows that $\vec{Y}' = \gamma \vec{Y}$. This implies
\[
\vec{M}' = \frac{1}{2}\args{\gamma \vec{X} + \gamma \vec{Y}} = \gamma \vec{M}
\]
showing that $\vec{M}$ and $\vec{M}'$ represent the same graph, that is $M = M'$. This contradicts our assumption and shows the assertion.
\qed

\subsection{Proof of Corollary \ref{corollary:midpoint}}\label{proof:corollary:midpoint}
A graph edit kernel space is a geodesic space \cite{Jain2015}, Theorem 3.3. This proves the first property. 

To show the second property, we assume that $M$ is a midpoint of $X$ and $Y$. By definition of a midpoint, we have 
\begin{align}\label{eq:corollary:midpoint:01}
\delta(X, M) + \delta(Y, M) = \delta(X, Y).
\end{align}
Since $\delta$ is a metric, the triangle inequality holds. We have
\begin{align}\label{eq:corollary:midpoint:02}
\delta(X, Y) \leq \delta(X, Z) + \delta(Z, Y).
\end{align}
for all $Z \in \S{G_A}$. Observe that the right hand side of Equation \eqref{eq:corollary:midpoint:02} is the sample Fr\'{e}chet function $F_2(Z)$ of $X$ and $Y$. From Equation \eqref{eq:corollary:midpoint:01} follows that a midpoint $M$ minimizes $F_2(Z)$ and therefore $M$ is a sample mean of $X$ and $Y$. 

Finally, the third property also follows from applying the triangle inequality. Suppose that $M$ is a sample mean. Then $M$ minimizes the right hand side of Equation \eqref{eq:corollary:midpoint:02}. Since a midpoint of $X$ and $Y$ exists, a minimizer satisfies Equation \eqref{eq:corollary:midpoint:01}. Thus, $M$ is a midpoint.
\qed

\subsection{Proof of Theorem \ref{theorem:convergence}}\label{proof:theorem:convergence}

For the proof, we invoke Zangwill's Convergence Theorem.

\subsection*{Zangwill's Convergence Theorem}

To state Zangwill's Convergence Theorem \cite{Zangwill1969}, we need to introduce some concepts. We assume that $\S{X}$ is a set and $2^{\S{X}}$ is the set of all subsets of $\S{X}$.

\begin{definition} A point-to-set mapping of $\S{X}$ is a map $\Phi: \S{X} \rightarrow 2^{\S{X}}$ that assigns each point $x \in \S{X}$ a subset $\Phi(x) \subset \S{X}$. 
\end{definition}

\begin{definition}
A point-to-set mapping $\Phi: \S{X} \rightarrow 2^{\S{X}}$ is said to be closed at $x\in \S{X}$ if the assumptions
\begin{enumerate}
\item $\lim_{k \to \infty} x_k = x$, $x_k \in \S{X}$
\item $\lim_{k \to \infty} y_k = y$, $y_k \in \Phi(x_k)$
\end{enumerate}
imply $y \in \Phi(x)$.
\end{definition}

\begin{theorem}[Zangwill's Convergence Theorem]\label{theorem:zangwill}
By $\S{X}_* \subseteq \S{X}$ we denote a solution set. Let $\Phi: \S{X} \rightarrow 2^{\S{X}}$ be a point-to-set map. For a given point $x_0 \in \S{X}$, let $\args{x_k}_{k \geq 0}$ be a sequence satisfying
\[
x_{k+1} \in \Phi(x_k)
\]
for all $k$. Suppose that
\begin{enumerate}
\item All points $x_k$ are contained in a compact subset of $\S{X}$.
\item There is a continuous function $f:\S{X} \rightarrow \R$ such that
\begin{enumerate}
\item $x \notin \S{X}_*$, then $f(y) < f(x)$ for all $y \in \Phi(x)$
\item $x \in \S{X}_*$, then $f(y) \leq f(x)$ for all $y \in \Phi(x)$
\end{enumerate}
\item The mapping $\Phi$ is closed at all points $\S{X} \setminus \S{X}_*$.
\end{enumerate}
Then the limit of any convergent subsequence of $\args{x_k}_{k \geq 0}$ is an element of $\S{X}_*$.
\end{theorem}

\subsubsection*{Proof}
Let $\S{X}$ be the set of all matrices that represent graphs from $\S{G_A}$. By $\pi:\S{X} \rightarrow \S{G_A}$ we denote the projection that maps each matrix $\vec{X}$ to the graph $X = \bracket{\vec{X}}$. Consider the set 
\[
\S{A}_n(\vec{X}) = \S{A}_1(\vec{X}) \times \cdots \times \S{A}_n(\vec{X}), 
\]
where $\S{A}_i(\vec{X})$ is the set of all representations $\vec{X}_{\!i} \in X_i$ optimally aligned to $\vec{X}$. To invoke Zangwill's Convergence Theorem, we first define the solution set
\[
\S{X}_* = \cbrace{\vec{M} = \frac{1}{n} \sum_{i=1}^n \vec{X}_{\!i} \,:\, \args{\vec{X}_1, \ldots, \vec{X}_n} \in \S{A}_n(\vec{M})}.
\]
By construction, the set $\pi(\S{X}_*)$ consists of all graph satisfying the necessary condition of optimality according to Theorem \ref{theorem:nesuco}.

The sample Fr\'echet function $F_n(Z)$ takes on the role of function $f(x)$ in Theorem \ref{theorem:zangwill}, item (2). We need to show that $F_n(Z)$ is continuous and satisfies properties (2a) and (2b) of Zangwill's Convergence Theorem. To show continuity, we rewrite the sample Fr\'echet function into the equivalent form
\[
F_n: \S{X} \rightarrow \R, \quad \vec{Z} \mapsto \sum_{i=1}^n \min_{\vec{X}_{\!i} \in X_i} \normS{\vec{X}_{\!i} - \vec{Z}}{^2}.
\]
The squared Euclidean distance is continuous. The minimum and the sum of a finite set of continuous functions is again continuous. This shows that $F_n$ is continuous. 

Next, we show properties (2a) and (2b). Let $\vec{Z} \in \S{X}$. We show that $F_n(\vec{M}) \leq F_n(\vec{Z})$ for every $\vec{M} \in \Phi(\vec{Z})$, where $\Phi$ is the point-to-set map determined by the MM-Algorithm. The set $\Phi(\vec{Z})$ is of the form
\[
\Phi(\vec{Z}) = \cbrace{\vec{M} = \frac{1}{n} \sum_{i=1}^n \vec{X}_{\!i} \,:\, \args{\vec{X}_1, \ldots, \vec{X}_n} \in \S{A}_n(\vec{Z})}.
\]
Let $\vec{M}\in \Phi(\vec{Z})$ be an arbitrary element. Then there are matrices $\vec{X}_i^{\vec{Z}} \in \S{A}_i(\vec{Z})$ such that 
\[
\vec{M} = \frac{1}{n} \sum_{i=1}^n \vec{X}_i^{\vec{Z}}.
\]
Then $\vec{M}$ is the unique minimum of the differentiable convex function 
\[
f_n(\vec{Z}) = \sum_{i=1}^n \normS{\vec{X}_i^{\vec{Z}} - \vec{Z}}{^2}.
\]
Thus, we have 
\begin{align}\label{eq:proof:theorem:convergence:eq1}
f_n(\vec{M}) \leq f_n(\vec{Z}) = F_n(\vec{Z}),
\end{align}
where strict inequality holds when $\vec{Z} \notin \S{X}_*$. Since $\vec{X}_i^{\vec{Z}}$ is optimally aligned to $\vec{Z}$ but not necessarily to $\vec{M}$, we have
\begin{align}\label{eq:proof:theorem:convergence:eq2}
F_n(\vec{M}) = \sum_{i=1}^n \min_{\vec{X}_{\!i} \in X_i} \normS{\vec{X}_{\!i} - \vec{M}}{^2} \leq \sum_{i=1}^n \normS{\vec{X}_i^{\vec{Z}} - \vec{Z}}{^2} = f_n(\vec{Z}).
\end{align}
Combining Equations \eqref{eq:proof:theorem:convergence:eq1} and \eqref{eq:proof:theorem:convergence:eq2} gives $F_n(\vec{M}) \leq F_n(\vec{Z})$ with strict inequality for $\vec{Z} \notin \S{X}_*$. This shows properties (2a) and (2b).

It remains to show that $\phi$ is a closed map. We first show that $\phi_X(\vec{Z}) = \S{A}_{X}(\vec{Z})$ is closed for all $\vec{Z} \in \S{X}$, where $\S{A}_{X}(\vec{Z})$ is the set of all optimal alignments between representations from graph $X$ and $\vec{Z}$. Let $\vec{Z}_k \to \vec{Z}$ and $\vec{X}_{\!k} \to \vec{X}$ convergent sequences with $\vec{X}_{\!k} \in \S{A}_X(\vec{Z}_k)$. From $\vec{X}_{\!k} \in \S{A}_X(\vec{Z}_k)$ follows $\delta(Z_k, X_k) = \norm{\vec{Z}_k - \vec{X}_{\!k} }$, where $Z_k = \pi(\vec{Z}_k)$ and $X_k = \pi(\vec{X}_{\!k} )$. 
Since the graph edit kernel metric is continuous, we have
\[
\lim_{k \to \infty} \delta\args{Z_k, X_k} = \lim_{k \to \infty} \norm{\vec{Z}_k - \vec{X}_{\!k} } = \norm{\vec{Z} - \vec{X}} = \delta(Z, X).
\]
From $\vec{X} \in \S{A}_X(\vec{Z})$ follows that the point-to-set map $\phi_X(\vec{Z})$ is closed. In a similar way we can show that $\phi_n(\vec{Z}) = \S{A}_n(\vec{Z})$ is closed. The function 
\[
\psi(\vec{X}_1, \ldots, \vec{X}_n) = \frac{1}{n}\sum_{i=1}^n \vec{X}_i
\]
is continuous and therefore closed. Then the composition $\Phi = \psi \circ \phi_n$ is a closed map. All assumptions of Zangwill's Convergence Theorem hold. This shows the first assertion. 

To show the second assertion, observe that the sequence $\args{v_k}_{k\geq 0}$ with $v_k = F_n(M_k)$ is monotonously decreasing and bounded from below by $0$. By the monotone convergence theorem, the sequence $\args{v_k}_{k\geq 0}$ is convergent and converges to its infimum $v_*$. 
\end{appendix}
\end{small}

\end{document}